\documentclass[10pt,twocolumn,letterpaper]{article}

\usepackage[pagenumbers]{cvpr} 
\usepackage{graphicx}
\usepackage{amsmath}
\usepackage{amssymb}
\usepackage{booktabs}
\usepackage{graphics}
\usepackage{microtype}
\usepackage[numbers,sort,compress]{natbib}
\usepackage{multirow}
\usepackage[usenames,dvipsnames]{xcolor}
\usepackage{adjustbox}
\usepackage{enumitem}
\usepackage{colortbl}
\usepackage{appendix}

\usepackage[pagebackref,breaklinks,colorlinks]{hyperref}

\usepackage[capitalize]{cleveref}
\crefname{section}{Sec.}{Secs.}
\Crefname{section}{Section}{Sections}
\Crefname{table}{Table}{Tables}
\crefname{table}{Tab.}{Tabs.}

\begin{document}

\title{HyperInverter: Improving StyleGAN Inversion via Hypernetwork}

\author{
Tan M. Dinh \hspace{10mm} Anh Tuan Tran \hspace{10mm} Rang Nguyen \hspace{10mm} Binh-Son Hua \vspace{3mm} \\ 
VinAI Research, Hanoi, Vietnam \vspace{2mm} 
}

\maketitle

\begin{abstract}

Real-world image manipulation has achieved fantastic progress in recent years as a result of the exploration and utilization of GAN latent spaces. GAN inversion is the first step in this pipeline, which aims to map the real image to the latent code faithfully. Unfortunately, the majority of existing GAN inversion methods fail to meet at least one of the three requirements listed below: high reconstruction quality, editability, and fast inference. We present a novel two-phase strategy in this research that fits all requirements at the same time. In the first phase, we train an encoder to map the input image to StyleGAN2 $\mathcal{W}$-space, which was proven to have excellent editability but lower reconstruction quality. In the second phase, we supplement the reconstruction ability in the initial phase by leveraging a series of hypernetworks to recover the missing information during inversion. These two steps complement each other to yield high reconstruction quality thanks to the hypernetwork branch and excellent editability due to the inversion done in the $\mathcal{W}$-space. Our method is entirely encoder-based, resulting in extremely fast inference. Extensive experiments on two challenging datasets demonstrate the superiority of our method. \footnote{Project 
page:  \href{https://di-mi-ta.github.io/HyperInverter}{https://di-mi-ta.github.io/HyperInverter}}

\end{abstract}

\section{Introduction}
Generative adversarial networks (GANs)~\cite{goodfellow2014generative} in modern deep learning have allowed us to synthesize expressively realistic images, a trend that has continued to flourish in recent years. GANs can now be trained to generate images of high-resolution~\cite{karras2018progressive} with diverse styles~\cite{karras2019stylegan,Karras2019stylegan2,karras2020ada} and apparently fewer artifacts~\cite{karras2021alias}. Moreover, the latent spaces learned by these models also encode a diverse set of interpretable semantics. These semantics provide a tool to manipulate the synthesized images. Therefore, understanding and exploring a well-trained GAN model is an important and active research area. Several studies~\cite{shen2020interpreting,yang2019semantic,shen2021closedform,harkonen2020ganspace,bau2020units} have been conducted to examine the latent spaces learned by GANs, which convey a wide range of interpretable semantics. 

\begin{figure}[t]
\centering
\includegraphics[width=0.85\linewidth]{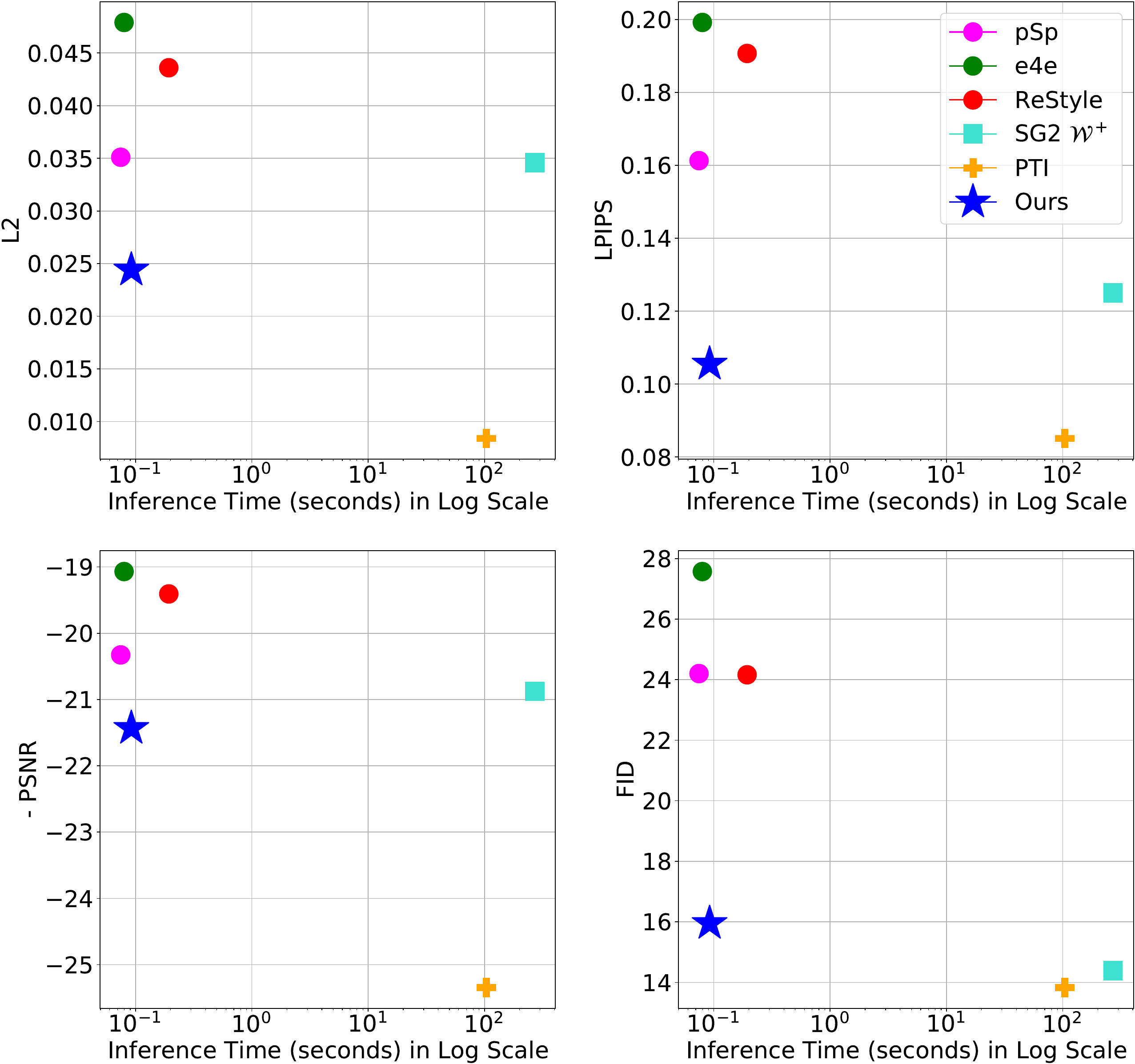}
\caption{Our end-to-end encoder-based method has more accurate reconstruction while having fast inference (toward the bottom left of each plot). As can be seen, our work outperform other encoder-based inversion methods (pSp~\cite{richardson2021encoding}, e4e~\cite{tov2021designing}, ReStyle~\cite{alaluf2021restyle}) significantly. Comparing with per-image optimization inversion technique~\cite{abdal2019image2stylegan}, our method is on par of quality but $3000\times$ faster. Only PTI~\cite{roich2021pivotal} has reconstruction quality better than our work. However, PTI requires per-image generator fine-tuning during the inference phase, which took a long time, $1100\times$ longer.}
\label{fig:teaser}
\end{figure}

To apply the semantic directions explored from GAN latent space to real-world images, the common-used practice is the "invert first, edit later" pipeline. \emph{GAN Inversion} is a typical line of work that aims to first map a real photograph to a latent code of a GAN model so that the model can accurately reconstruct the photograph. Then, we can manipulate the latent code to edit different attributes of the reconstructed picture. There are two general approaches to determining a latent code of a GAN model given an input image: by an iterative optimization~\cite{creswell2018inverting, ma2019invertibility, abdal2019image2stylegan, Karras2019stylegan2, kang2021gan} and by inference with an encoder~\cite{zhu2016generative,pidhorskyi2020adversarial, richardson2021encoding, tov2021designing}. 
The optimization-based method tends to perform reconstruction more accurately than encoder-based one but requires significantly more computation time, hindering use cases for interactive editing. 

While the goal of GAN inversion is not only to reconstruct the input image faithfully but also to effectively perform image editing later, there is the so-called reconstruction-editing trade-off raised by multiple previous works \cite{zhu2020improved, tov2021designing}.
This trade-off is shown to depend on the embedding space where an input image is mapped to. The native StyleGAN $\mathcal{W}$ space and the extended version $\mathcal{W}^{+}$ space~\cite{abdal2019image2stylegan}  are two most popular embedding spaces for StyleGAN inversion. 
Specifically, inverting an image to $\mathcal{W}$ space usually has excellent editability, but they are proved to be infeasible to reconstruct the input image faithfully~\cite{abdal2019image2stylegan}. On the contrary, $\mathcal{W}^{+}$ space allows to obtain more accurate reconstructions but it surfers from editing ability~\cite{tov2021designing}. 
To mitigate the effects of this trade-off, a diverse set of methods has been proposed. Some of them~\cite{tov2021designing, zhu2020improved} introduce the ways (e.g., regularizer or adversarial training) to select the latent code in the high editable region of $\mathcal{W}^{+}$ space and accept a bit of sacrifice in the reconstruction quality. Another option is to utilize a two-stage approach. PIE~\cite{tewari2020pie} opts first to use an optimization process to locate the latent code in $\mathcal{W}$ space to preserve the editing ability and then optimize further the latent in the $\mathcal{W}^{+}$ space to enhance the reconstruction quality. PTI~\cite{roich2021pivotal} approaches the same as PIE in the first stage. However, in the second stage, they choose the generator fine-tuning option to improve reconstruction results further.
As can be seen, such methods require either optimization or fine-tuning process for each new image, leading to expensive inference time.

Motivated from above weakness, we propose a novel pure encoder-based two-phase method for StyleGAN inversion. Our method not only runs very fast but also obtains robust reconstruction quality. Specifically, in the first phase, we change to use a standard encoder to regress the image to the latent code in $\mathcal{W}$ space instead of utilizing optimization process as PTI and PIE. Then, in the second phase, we leverage the hypernetworks to predict the \emph{residual weights}, which can recover the lost details of the input image after the first phase. We then use the residual weights to update the original generator for synthesizing the final reconstructed image. Our design would help migrate optimization or fine-tuning procedures in the inference phase, significantly reducing processing time. 
In summary, our contributions are:
\begin{itemize}[leftmargin=*]
\item A completely encoder-based two-phase GAN inversion approach allows faithful reconstruction of real-world images while keeping editability and having fast inference;
\item A novel network architecture that consists of a series of hypernetworks to update the weights of a pre-trained StyleGAN generator, thereby improving reconstruction quality;
\item An extensive benchmark of GAN inversion that demonstrates the superior performance of our method compared to the existing works;
\item A new method for real-world image interpolation that interpolates both latent code and generator weights.
\end{itemize}

\section{Related Work}

\noindent\textbf{Latent Space Manipulation}  
The latent space of a well-trained GAN generator (e.g., StyleGAN) contains various interpretable semantics of real-world images, which provides a wonderful tool to perform a diverse set of semantic image editing tasks. Therefore, understanding and exploring the semantic directions encoded in the pre-trained GAN latent space have been the subject of numerous studies. Some researches~\cite{goetschalckx2019ganalyze, shen2020interpreting, yang2021discovering}  use complete supervision in the form of semantic labels. Such methods require either a well-trained attribute classifier or the images annotated with editing attributes. Therefore, this condition prevents the applications of supervised methods to a limited of known attributes. As a result, other studies~\cite{harkonen2020ganspace, voynov2020unsupervised, wang2020geometric, shen2021closedform} propose unsupervised approaches to accomplish the same aim without the need for manual annotations. These types of methods allow exploring various fancy editing directions that we did not know before. Besides, some researches~\cite{ren2021DisCo, Yuksel_2021_ICCV} also use the idea of contrastive learning to analyze the GAN latent space.

\noindent\textbf{GAN Inversion.} To apply such latent manipulations to the real-world image, we first need to locate the latent code represented for that image in the pre-trained GAN latent space. This process is known as \emph{GAN inversion}~\cite{zhu2016generative}. Existing GAN inversion methods can be categorized into two general groups: optimization-based and encoder-based approaches. Optimization-based methods~\cite{creswell2018inverting, ma2019invertibility, abdal2019image2stylegan, Karras2019stylegan2, kang2021gan} directly optimize the latent vector by minimizing the reconstruction error for each given image. These methods usually provide high-quality reconstructions but require too much time to perform, and it is tough to apply on real-time applications. Encoder-based works~\cite{zhu2016generative,pidhorskyi2020adversarial, richardson2021encoding, tov2021designing, wei2021simpleinversion} employ an encoder to learn a direct mapping from a given picture to its corresponding latent vector, which allows for fast inference. However, the reconstruction quality of encoder-based techniques is usually worse than that of optimization-based approaches. Therefore, some hybrid works~\cite{bau2019seeing, zhu2020domain, guan2020collaborative, bau2020semantic} have been proposed, which combine both above general approaches to partly balance the trade-off of reconstruction quality and inference time. They first use an encoder to encode the image to the initial latent code and then use this latent as an initial point for the later optimization process. Hybrid methods can be considered as the subset of the two-stage methods and they use both optimization and encoder techniques in the method. In addition, there are other two-stage approaches but are not hybrid ones. For example, PIE~\cite{tewari2020pie} uses optimization methods on both phases, while PTI~\cite{roich2021pivotal} combines optimization process with the generator fine-tuning technique. In comparison, our work is also a two-stage method but differs from the above ones in that our approach is only based purely on the encoder-based manner in both phases.

\begin{figure*}[!]
\centering
\includegraphics[width=\linewidth]{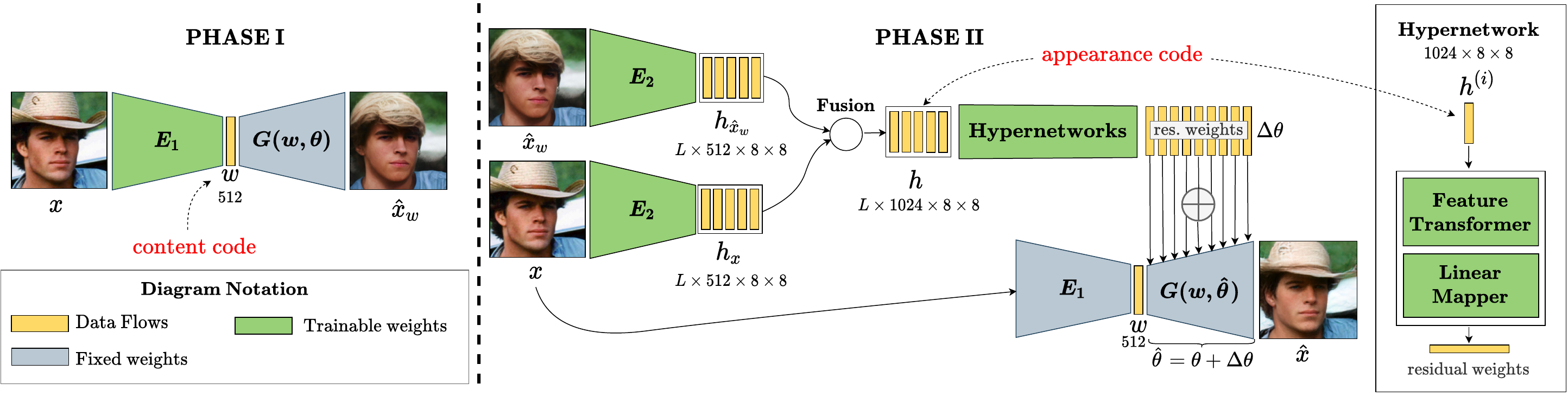}
\caption{Our method contains \textbf{two} sequential phases: (1) We first train an encoder $E_{1}$ to encode the input image $x$ to a \emph{content} code $w$ in $\mathcal{W}$-space; $w$ represents the main semantics of the image, therefore used in editing later. The output image of this phase is  $\hat{x}_{w}$. (2) We further regress the \emph{residual weights} to update the generator to faithfully reconstruct the input details. First, we use another encoder $E_{2}$ and a fusion operator to extract the \emph{appearance} code $h$ from the input image $x$ and the initial image $\hat{x}_{w}$, where $L$ is the number of style layers of StyleGAN. Then, we employ a series of hypernetworks to embed the appearance code $h$ to the generator $G$ by predicting the residual weights $\Delta\theta$. The final reconstructed image $\hat{x}$ is generated by $G$ with updated weights $\hat{\theta} = \theta + \Delta\theta$ and $w$ content code from Phase I.} 
\label{fig:method}
\end{figure*}

\noindent\textbf{Hypernetwork. } Hypernetwork is the auxiliary neural network that produces the weights for other network (often called as \emph{primary network}). They were first proposed by Ha et al.~\cite{ha2016hypernetworks} and have been used in a wide range of applications from semantic segmentation~\cite{nirkin2021hyperseg}, 3D scene representation~\cite{littwin2019deep, sitzmann2020implicit}, neural architecture search (NAS)~\cite{zhang2018graph} to continual learning~\cite{oshg2019hypercl}. In this study, we design the hypernetworks to improve the quality of our GAN inversion method. Specifically, we leverage hypernetworks to update the weights of the pre-trained GAN generator instead of fine-tuning procedure, which took a lot of processing time. Interestingly, a concurrent work named HyperStyle~\cite{alaluf2021hyperstyle} also leverages the hypernetworks to solve the StyleGAN inversion task at the same time as ours.

\section{Method}

Our method is an end-to-end encoder-based GAN inversion approach with two consecutive phases. The first phase is a standard encoder that first maps input image to the GAN latent space (Section \ref{subsec:phase_1}).
The second phase then refines the generator so that it adapts to the reconstructed latent code to preserve original image details while allowing editability (Section \ref{subsec:phase_2}).
Figure~\ref{fig:method} provides an overview of our approach. 
Let us now describe the details of each phase below.

\subsection{Phase I: From image to content code}
\label{subsec:phase_1}

In StyleGAN, $\mathcal{W}$ space is empirically proven to have excellent editing ability compared to the popular $\mathcal{W^{+}}$ space for inversion~\cite{zhu2020improved, tov2021designing}. Therefore, we first choose $\mathcal{W}$ space to locate our latent code. Specifically, given the input image $x$, we train an encoder $E_{1}$ to regress $x$ to the corresponding latent code $w \in \mathcal{W}$:
\begin{equation}
   w = E_{1}(x),
\end{equation}
where $w$ has the size of $\mathbb{R}^{512}$. This $w$ code has a role as the \emph{content} code encoding the main semantics in the image. For image editing task later, one could traversal this $w$ code on the interpretable semantic directions to manipulate the image. 
The reconstructed image $\hat{x}_{w}$ can be computed by passing the latent code $w$ to the generator: 
\begin{equation}
 \hat{x}_{w} = G(w, \theta),
\end{equation}
where $G$ is the well-trained StyleGAN generator with convolutional weights $\theta$. 

A typical problem of the aforementioned inversion is that the image $\hat{x}_w$ is perceptually different from the input $x$, making the editing results less convincing. This phenomenon is caused by the relatively low dimensionality of the latent code $w$, which makes it challenging to represent all features from the input image. 
To address this weakness, we propose a novel phase II focusing on improving further the reconstruction quality by refining generator $G$ with new weights.

\subsection{Phase II: Generator refinement via hypernetworks from appearance code.}
\label{subsec:phase_2}

Our goal in this phase is to recover the missing information of input $x$ and thus reduce the discrepancy between $x$ and $\hat{x}_w$. 
Other two-stage methods use either per-image optimization process~\cite{tewari2020pie} or per-image fine-tuning $G$~\cite{roich2021pivotal} to improve reconstruction quality further. However, the drawback of such these approaches is the slow inference time. We instead opt for a different approach that aims to create a single-forward pass network to learn to refine the generator $G$ by inspecting the current input $x$ and the reconstruction $\hat{x}_w$. Using single-forward pass network can guarantee the very fast running time. This network first encodes $x$ and $\hat{x}_w$ into intermediate features and fuses them into a common feature tensor that can be used by a set of hypernetworks to output a set of \emph{residual weights}. We then use the residual weights to update the corresponding convolutional layers in the generator $G$ to yield a new set of weights $\hat{\theta}$. It is expected that the refined reconstruction from $G(\cdot, \hat{\theta})$ should be closer to the input $x$ than the current reconstruction $\hat{x}_w$.

\begin{figure}[t]
\centering
\includegraphics[width=\linewidth]{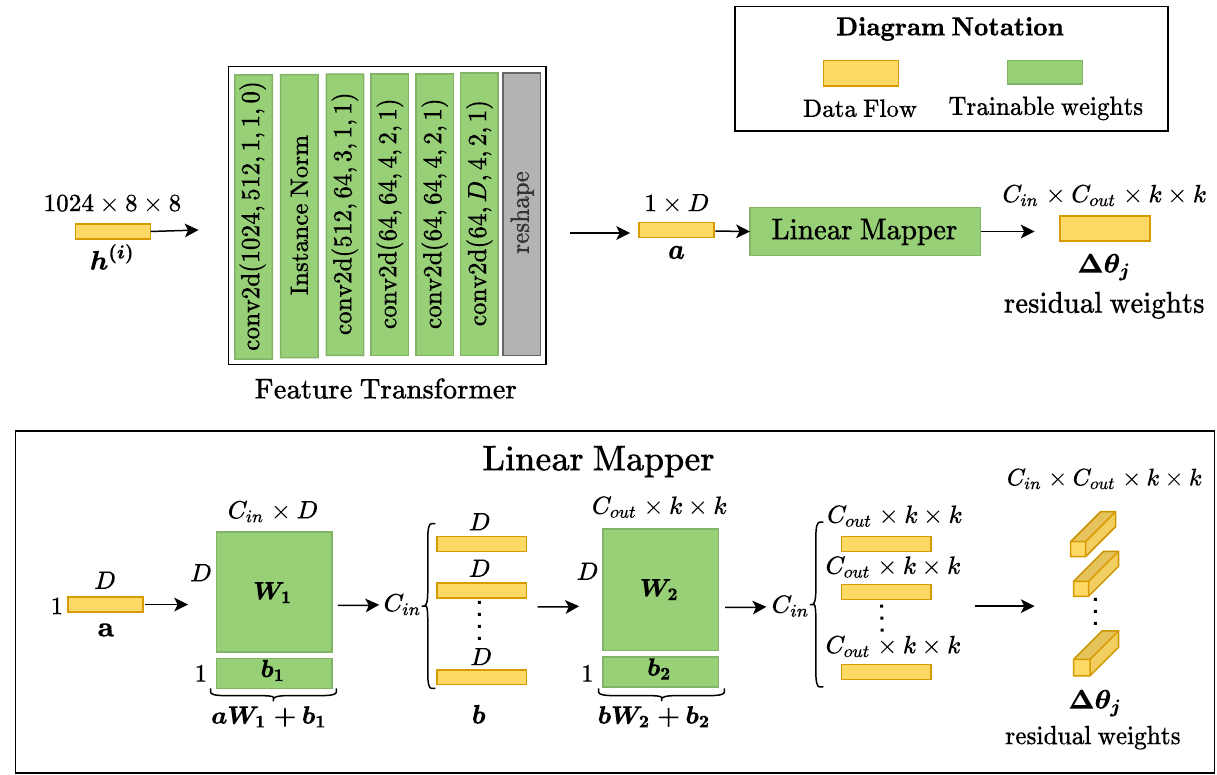}
\caption{\textbf{Hypernetwork design} for predicting the residual weights $\Delta \theta_{j}$ of the convolutional layer $j$ in the StyleGAN generator. Here we assume layer $j$ receives the style vector indexed by $i$. Diagram notation: conv2d(in\_channels, out\_channels, kernel\_size, stride, padding). For simplicity, we omit in the figure the ReLU~\cite{agarap2018deep} activation layer after each conv2d layer.}
\vspace{-0.1in}
\label{fig:hypernetwork}
\end{figure}

Particularly, we first use a shared encoder $E_{2}$ to transform both image $x$ and $\hat{x}_{w}$ into the intermediate features:
\begin{align}
h_{\hat{x}_{w}} &= E_2(\hat{x}_{w}), \\
h_{x} &= E_2(x), \nonumber
\end{align}
where $h_{\hat{x}_{w}}$, $h_{x} \in \mathbb{R}^{L \times 512 \times 8 \times 8}$, and $L$ is the number of style layers from the pretrained StyleGAN generator. Recall that in StyleGAN generator network, each convolutional layer receives an input that corresponds to one of $L$ style vectors. Therefore, we choose to regress to $L$ feature components rather than a common one motivated by the StyleGAN generator's design. Then we use a fusion operator to combine these two features forming an appearance code $h$, with $h \in \mathbb{R}^{L \times 1024 \times 8 \times 8}$.

To embed the appearance code $h$ to the reconstructed image, we opt to use $h$ to update the weights of the pretrained StyleGAN generator $G$. Motivated by the hypernetwork's idea~\cite{ha2016hypernetworks}, which use an small extra neural network to predict the weights for the primary network, we leverage a series of hypernetworks to predict the residual weights, which then are added to refine the weights of $G$. 
Here we only consider predicting the weights for the main convolutional layers of $G$, skipping biases and other layers. 

Assume that the pretrained generator $G$ has $N$ convolutional layers with weights $\theta = (\theta_{1}, \theta_{2}, ..., \theta_{N})$. 
In StyleGAN architecture, each convolution layer $j \in \{1..N\}$ receives a corresponding style vector indexed by $i\in \{1..L\}$ as input where $L$ is the total number of style vectors. 
We therefore propose to use a small hypernetwork $H_{j}$ to predict the residual weights $\Delta \theta_{j}$ of each convolutional layer $j$, which results in a total of $N$ hypernetworks. The residual weights $\Delta \theta_{j}$ for convolutional layer $j$ is computed as 
\begin{equation}
    \Delta \theta_{j} = H_{j}(h^{(i)}).
\end{equation}
The details of the design of hypernetwork is illustrated in Figure~\ref{fig:hypernetwork}. Specifically, each hypernetwork contains two main modules, which are feature transformer and linear mapper. 
The feature transformer is a small convolutional neural network to transform the appearance code $h^{(i)}$ to the hidden features. 
The linear mapper is used for the final mapping from the hidden features to the convolutional weights. We adopt the technique proposed by Ha et al.~\cite{ha2016hypernetworks} that uses two small matrices instead of a single large matrix to the weight mapping to keep the number of parameters in the hypernetwork manageable. 

Finally, the updated generator has the convolutional weights as $\hat{\theta} = ( \theta_{1} + \Delta \theta_{1}, \theta_{2}+ \Delta \theta_{2}, ..., \theta_{N}+ \Delta \theta_{N})$.
The final reconstructed image $\hat{x}$ could be obtained from the updated generator and the content code $w$ from Phase I:
\begin{equation}
\hat{x} = G(w, \hat{\theta}).
\end{equation}

\subsection{Loss functions}
We train our network in the two phases independently. In Phase I, we employ a set of loss functions to ensure faithful reconstruction.
Given $x$ as the input image, $\hat{x}$ is the reconstructed image, we define our reconstruction loss $\mathcal{L}_{rec}$ as a weighted sum of
\begin{align}
\mathcal{L}_{rec} = \lambda_{pixel}\mathcal{L}_{2} + \lambda_{perc}\mathcal{L}_{LPIPS} + \lambda_{id}\mathcal{L}_{ID}, 
\end{align}
where $\lambda_{pixel}$, $\lambda_{perc}$, $\lambda_{id}$ are the hyper-parameters. Each loss is defined as follows:
\begin{itemize}
\item $\mathcal{L}_{2} = || x - \hat{x} ||_{2}$ measures the pixel-wise similarity,

\item $\mathcal{L}_{LPIPS} = ||\mathcal{F}_{LP}(x) - \mathcal{F}_{LP}(\hat{x})||_{2}$ is the perceptual loss~\cite{zhang2018unreasonable}, where $\mathcal{F}_{LP}$ indicates a perceptual feature extractor~\cite{zhang2018unreasonable}. We use the pre-trained AlexNet~\cite{krizhevsky2012imagenet} version, similar to  previous GAN inversion works. 

\item $\mathcal{L}_{ID} = 1 - \langle \mathcal{F}_{ID}(x), \mathcal{F}_{ID}(\hat{x}) \rangle$, where $\mathcal{F}_{ID}$ is a class-specific feature extractor network, which is a pre-trained ArcFace~\cite{deng2019arcface} for human facial domain or a ResNet-50~\cite{he2016deep} pre-trained with MOCOv2~\cite{chen2020improved} for Churches domain, $\langle . \rangle$ denotes cosine similarity between two feature embedding vectors.
\end{itemize}

In Phase II, we further use the non-saturating GAN loss~\cite{goodfellow2014generative} in addition to the reconstruction loss in Phase I. This loss helps to ensure the realism of generated images since we modify the weights of the generator in this phase: 
\begin{align}
    \mathcal{L}_{enc}= \mathcal{L}_{rec} + \lambda_{adv}\mathcal{L}_{adv}
\end{align}
where
\begin{align}
\mathcal{L}_{adv} = - \mathbb{E}_{\hat{x} \sim p_{\hat{X}}}[\log(D(\hat{x}))], 
\end{align}
 $D$ is initialized with the weights from the well-trained StyleGAN discriminator, $\lambda_{adv}$ is the hyperparameter balancing two losses.
Discriminator $D$ is trained along with our network in a adversarial manner. We further impose $\mathcal{R}_{1}$ regularization~\cite{mescheder2018training} to $D$ loss. The final loss for $D$ is: 
\begin{align}
\mathcal{L}_{D} = -\mathbb{E}_{x \sim p_{X}}[\log(D(x))] - \mathbb{E}_{\hat{x} \sim p_{\hat{X}}}[\log(1 - D(\hat{x}))] \\
\nonumber  + \dfrac{\gamma}{2} \mathbb{E}_{x \sim p_{X}}[||\nabla_{x} D(x)||_{2}^{2}].
\end{align}

\section{Experiments}

\subsection{Experimental Settings}
\noindent\textbf{Datasets.} We conduct the experiments on two challenging domains, which are \emph{human faces} and \emph{churches}. Those are selected because reconstructing human faces is a popular task in GAN inversion while reconstructing churches is a common test for generating outdoor scene images. For human facial domain, we employ the FFHQ~\cite{karras2019stylegan} dataset as our training set, and the official test set of CelebA-HQ~\cite{liu2015deep, karras2018progressive} as our test set. For churches domain, we use the LSUN Church~\cite{yu2015lsun} dataset. We adopt the official train/test split of LSUN Church for our training and testing sets, respectively.

\noindent\textbf{Baselines.} We compare our method with various GAN inversion approaches including optimization-based, encoder-based, and two-stage methods. For optimization-based works, we compare our method with the inversion technique proposed by~\cite{abdal2019image2stylegan}, denoted as SG2 $\mathcal{W}^{+}$. For encoder-based approaches, we choose pSp~\cite{richardson2021encoding}, e4e~\cite{tov2021designing}, and ReStyle~\cite{alaluf2021restyle} (apply over e4e backbone) to compare with our results. For other two-stage methods, we compare with PTI~\cite{roich2021pivotal}. We use the official pre-trained weights and configurations released from the authors to perform our evaluation experiments.

\noindent\textbf{Implementation Details.} In our experiments, the pre-trained StyleGAN generators and discriminators being used are obtained directly from StyleGAN2~\cite{Karras2019stylegan2} repository.  
To implement the encoders $E_1$ and $E_2$, we adopt the backbone design of ~\cite{richardson2021encoding, tov2021designing}. For $E_2$, we modify the style block of original backbone to output $512 \times 8 \times 8$ tensors instead of $512$ vectors. In Phase I, we follow the previous encoder-based methods~\cite{richardson2021encoding, tov2021designing, alaluf2021restyle} and use the Ranger optimizer, which combines the  Lookahead~\cite{NEURIPS2019_90fd4f88} and the Rectified Adam~\cite{liu2019variance} optimizer, for training. 
In Phase II, we use Adam~\cite{kingma2014adam} with standard settings, which we found to make the training of the hypernetworks converges faster. 
In both phases, we set the learning rate to a constant of $0.0001$. 
For hyperparameters in the loss functions, we set $\lambda_{pixel}=1.0$ and $\lambda_{perc}=0.8$ for both domains. 
For other hyperparameters, we use $\lambda_{id}=0.1$, $\lambda_{adv}=0.005$, $\gamma=10$ for the human facial domain, and $\lambda_{id}=0.5$, $\lambda_{adv}=0.15$, $\gamma=100$ for the churches domain. Empirically, we found that applying adversarial loss and $\mathcal{R}_{1}$ regularization later in the training process leads to better stability than using them from the beginning. Specifically, we first train the model with batch size $8$ for the first $200,000$ iterations. Then, we add the adversarial loss and $\mathcal{R}_{1}$ regularization to the training process and continue the training with batch size $4$ until convergence. All evaluation experiments are conducted using a single NVIDIA Tesla V100 GPU.

\begin{figure}[h]
\centering
\includegraphics[width=\linewidth]{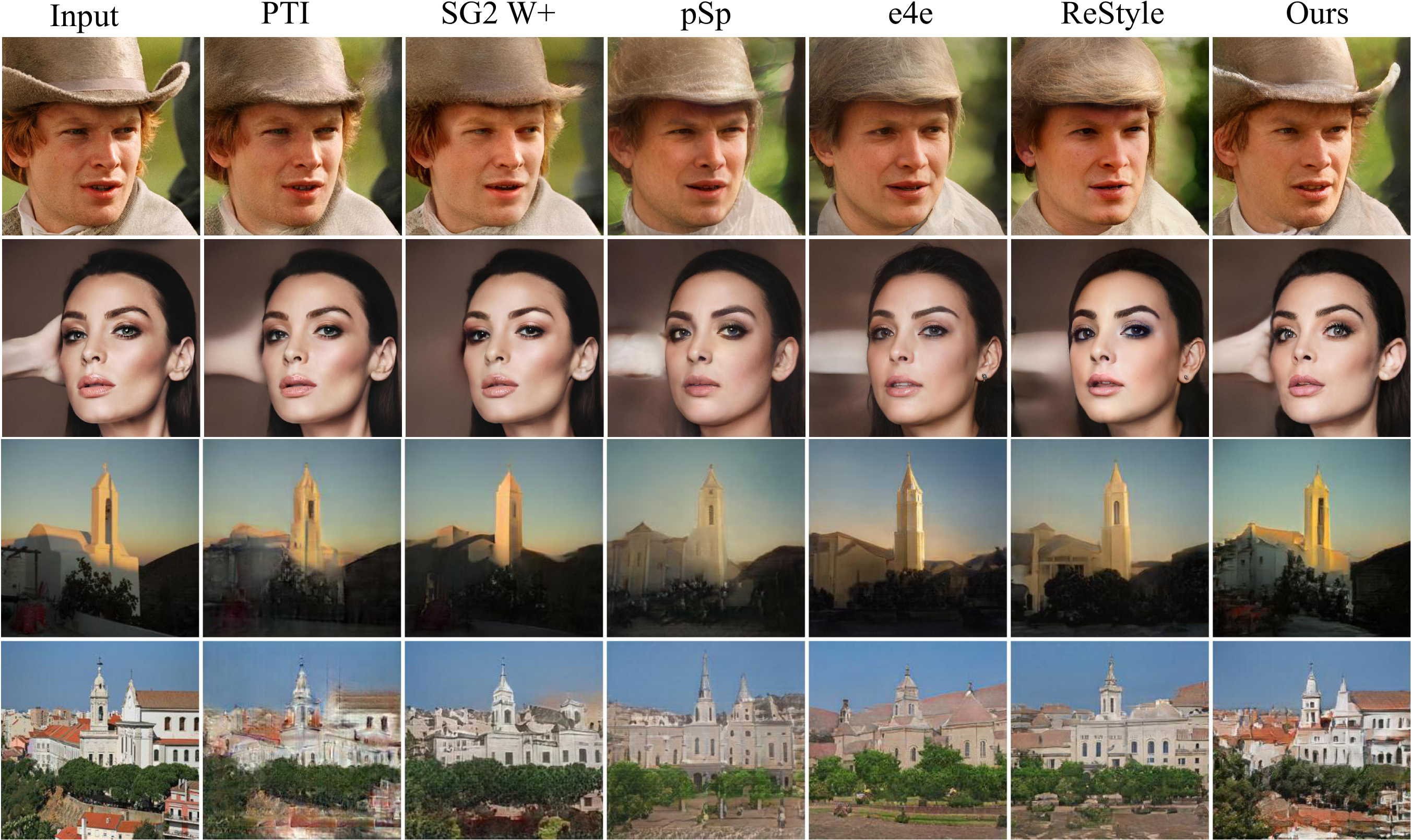}
\caption{\textbf{Qualitative reconstruction comparison} of our method with the current state-of-the-art StyleGAN inversion approaches. 
More examples are in supplementary. Best viewed in zoom.}
\label{fig:reconstruction_comparison}
\vspace{-4mm}
\end{figure}

\begin{table*}
\begin{center}
\resizebox{0.95\linewidth}{!}{%
\begin{tabular}{l | l | c c c c c c c | c } 
\toprule
Domain & Method & L2 ($\downarrow$)  & LPIPS ($\downarrow$) & ID ($\uparrow$) & FID ($\downarrow$) &  KID\textsuperscript{($\times 10^{3}$)} ($\downarrow$) & PSNR ($\uparrow$) & MS-SSIM ($\uparrow$) & Time (s) ($\downarrow$)\\
\midrule
\multirow{6}{*}{\stackbox{Human\\Faces}} & SG2 $\mathcal{W}^{+}$~\cite{abdal2019image2stylegan} & \textbf{\textcolor{red}{0.0346}} & \textbf{\textcolor{red}{0.1250}} & 0.6669 & 14.39 & 5.41 & \textbf{\textcolor{red}{20.8771}} & 0.7181 & 270 \\

&  PTI~\cite{roich2021pivotal}  & 0.0084 & 0.0851 & 0.8387 & 13.83 & 5.06 & 25.3447 & 0.7799 & 104 \\

\cmidrule{2-10}
                             & pSp~\cite{richardson2021encoding} & \underline{0.0351} & \underline{0.1613} & \underline{0.5596} & 24.20 & 12.13 & \underline{20.3277} & \underline{0.6496} & 0.0746 \\
                             & e4e~\cite{tov2021designing} & 0.0479 & 0.1992 & 0.4963 & 27.57 & 13.90 & 19.0703 & 0.6231 & 0.0797 \\
                             & ReStyle~\cite{alaluf2021restyle} & 0.0436 & 0.1907 & 0.5036 & \underline{24.16} & \underline{10.80} & 19.4076 & 0.6309 & 0.1932 \\
                             & \cellcolor{yellow!15} Ours & \cellcolor{yellow!15} \textbf{0.0243} & \cellcolor{yellow!15} \textbf{0.1054} & \cellcolor{yellow!15} \textbf{0.6013} & \cellcolor{yellow!15} \textbf{15.88} & \cellcolor{yellow!15} \textbf{5.63} & \cellcolor{yellow!15} \textbf{21.4417} & \cellcolor{yellow!15}\textbf{0.6725} & \cellcolor{yellow!15} 0.0920 \\
\midrule
\multirow{5}{*}{Churches} & SG2 $\mathcal{W}^{+}$~\cite{abdal2019image2stylegan} & \textbf{\textcolor{red}{0.1668}} & \textbf{\textcolor{red}{0.3250}} & / & \textbf{\textcolor{red}{50.00}} & \textbf{\textcolor{red}{7.32}} & \textbf{\textcolor{red}{14.8427}} & \textbf{\textcolor{red}{0.4797}} & 181 \\
&  PTI~\cite{roich2021pivotal}  & 0.0506 & 0.0971 & / & \textbf{\textcolor{blue}{76.21}} & \textbf{\textcolor{blue}{28.29}} & 19.0105 & 0.6968 & 69 \\

\cmidrule{2-10}
                             & pSp~\cite{richardson2021encoding} & \underline{0.1107} & \underline{0.3590} & / & 57.42 & 14.35 & \underline{15.7509} & \underline{0.4070} & 0.0521 \\
                             & e4e~\cite{tov2021designing} & 0.1414 & 0.4209 & / & 55.40 & 11.03 & 14.7071
 & 0.3481 & 0.0534 \\
                             & ReStyle~\cite{alaluf2021restyle} & 0.1272 & 0.3775
 & / & \underline{52.16} & \underline{9.04} & 15.1849 & 0.3878 & 0.1028 \\
                             & \cellcolor{yellow!10} Ours & \cellcolor{yellow!15} \textbf{0.0910} & \cellcolor{yellow!15} \textbf{0.2226} & \cellcolor{yellow!15} / & \cellcolor{yellow!15} \textbf{46.96} & \cellcolor{yellow!15} \textbf{6.82} & \cellcolor{yellow!15} \textbf{16.7152} & \cellcolor{yellow!15} \textbf{0.5762} & \cellcolor{yellow!15} 0.0661 \\

\bottomrule
\end{tabular}
}
\end{center}
\vspace{-0.15in}
\caption{\textbf{Quantitative reconstruction results with inference time} of our method compared to the state-of-the-art StyleGAN inversion approaches. The \textbf{best} and \underline{runner-up} results within \emph{encoder-based methods} are marked in bold and underline, respectively. Values in \textbf{\textcolor{blue}{blue}} and \textbf{\textcolor{red}{red}} highlight the cases that we outperform PTI~\cite{roich2021pivotal} and SG2 $\mathcal{W}^{+}$~\cite{abdal2019image2stylegan}, respectively.}
\label{tab:reconstruction_quatitative_results}
\end{table*}

\subsection{Reconstruction Results}

\noindent\textbf{Quantitative Results.} We use a diverse set of metrics to measure the reconstruction quality of our method compared with existing approaches. Specifically, we use the pixel-wise L2,  perceptual LPIPS~\cite{zhang2018unreasonable}, MS-SSIM~\cite{wang2003multiscale}, and PSNR metrics. We also evaluate the realisticity of reconstructed images by using  KID~\cite{binkowski2018demystifying} and FID~\cite{heusel2017gans} metrics. For the human facial domain, we measure the ability to preserve the subject identity of each inversion method by computing the identity similarity between the source images and the reconstructed ones using an off-the-shelf face recognition model CurricularFace~\cite{huang2020curricularface}. Besides,  we also consider the \emph{quality-time trade-off} raised by \cite{alaluf2021restyle} in our evaluation pipeline by reporting the inference time of each approach. 

The results can be found in Table~\ref{tab:reconstruction_quatitative_results}. Our method significantly outperforms other encoder-based methods for reconstruction quality on both domains. Compared to such methods using optimization (and fine-tuning) technique, we have comparable performance. On the human facial domain, we win against $\text{SG2}~\mathcal{W}+$ on L2, LPIPS, and PSNR. On the churches domain, we significantly outmatchs other method except PTI on all metrics. PTI still surfers from distortion-perception trade-off, exposed by the worst FID and KID. Overall, while our method does not outperform optimization-based methods completely, it is $3000 
\times$ and $1100\times$ faster than $\text{SG2}~\mathcal{W}+$ and PTI, respectively. 
Further reducing the performance gap for encoder-based and optimization-based (with generator fine-tuning) methods is left for future work. 

\noindent\textbf{Qualitative Results}
We visualize the reconstructions in Figure~\ref{fig:reconstruction_comparison}. 
For faces, we found that our method is particularly robust at preserving details of the accessories, e.g., see the cowboy hat example, and the background, e.g., see the hand example.
These curated results also highlight that optimization-based methods do not perform well in such cases (despite they are better overall, which is reflected in the quantitative results in Table~\ref{tab:reconstruction_quatitative_results}).
In churches domain, our method reconstructed images significantly better than other methods including both PTI and $\text{SG2}~\mathcal{W}+$ in terms of both distortion and perception. Since PTI suffers from the distortion-perception trade-off as mentioned above, so the pictures from PTI are not completely realistic. Additional qualitative results can be found in the supplementary. 

\subsection{Editing Results}

\noindent\textbf{Quantitative Results.} We first perform a quantitative evaluation for editing ability. Following previous works~\cite{zhu2020improved, roich2021pivotal}, we present an experiment to test the effect of editing operator with the same editing magnitude on the latent code inverted by different inversion methods. We opt for age and rotation for two editing directions in this experiment. 
Given the latent code $w$, we apply the editing operator to obtain the new latent code as $w\textsubscript{edit} = w + \gamma * d$ where $\gamma$ is the editing magnitude and $d$ is the semantic direction learned by InterFaceGAN~\cite{shen2020interpreting}.
To quantitatively evaluate the editing ability, we measure the amount of age change for age edit and yaw angle change (in degree) for pose edit when applying the same $\gamma$ on each baseline. 
We employ the DEX VGG~\cite{rothe2018deep} model for age regression, and a pre-trained FacePoseNet \cite{chang17fpn} for head pose estimation. 
The results are shown in Table~\ref{tab:editing_evaluation}. As can be seen, as we expected, the inversion done by our Phase I $\mathcal{W}$ encoder achieves the most significant effects since it works on highly editable $\mathcal{W}$-space latent code. 
After Phase 2 refinement, our method still preserves the capability of editing, outperforming all previous methods despite not being as good as the results in Phase 1.

\begin{table}
\begin{center}
\resizebox{\columnwidth}{!}{%
\begin{tabular}{l | r | r r r r r r} 
\toprule
 Dir. & $\gamma$ & e4e & ReStyle &  SG2 $\mathcal{W}^{+}$  & PTI & \stackbox{\textbf{Ours}\\($\mathcal{W}$ enc.)} & \stackbox{\textbf{Ours}\\ (full)} \\
\midrule
\multirow{8}{*}{\rotatebox{90}{Age}} & -3  &  -9.51 & -4.55 & -5.45 & -11.17 & \textbf{-22.69} & \underline{-20.58} \\
                     & -2  &  -5.6 & -2.78 & -3.79 & -6.71 & \textbf{-15.96} & \underline{-12.01} \\
                     & -1  &  -2.44 & -1.2 & -1.92 & -3.15 & \textbf{-5.51} & \underline{-4.56} \\
                     & 1   &  2.72 & 1.24 & 2.17 & 2.68 & \textbf{6.55} & \underline{4.78} \\
                     & 2   &  6.82 & 3.47 & 4.15 & 6.08 & \textbf{13.04} & \underline{9.8} \\
                     & 3   &  11.64 & 5.61 & 6.79 & 10.3 & \textbf{20.41} & \underline{14.25} \\
\midrule

\multirow{8}{*}{\rotatebox{90}{Pose}} & -3  &  7.68 & 5.14 & 5.79 & 8.50 & \textbf{13.63} & \underline{11.56}  \\
                      & -2  & 5.21 & 3.57  & 3.99  &  5.96 & \textbf{9.01} & \underline{7.73}  \\
                      & -1  & 2.75 & 1.85  & 2.00  & 3.15 & \textbf{4.64} & \underline{4.01} \\
                      & 1   & -2.75 & -1.77 & -1.90  & -2.95 & \textbf{-4.99} & \underline{-4.17} \\
                      & 2   & -5.36 & -3.50 & -3.83 & -5.89 & \textbf{-9.83} & \underline{-8.39} \\
                      & 3   & -7.96 & -5.14 & -5.70 & -8.88 & \textbf{-15.06} & \underline{-12.93}  \\
\bottomrule
\end{tabular}
}
\end{center}
\vspace{-4mm}
\caption{\textbf{Quantitative evaluation of editability.} We measure the amount of age change for age edit and yaw angle change (in degree) for pose edit when applying the same editing magnitude $\gamma$ on each method. The \textbf{best} and \underline{runner-up} values are marked in bold and underline, respectively.}
\vspace{-4mm}
\label{tab:editing_evaluation}
\end{table}

\begin{figure*}[t]
\centering
\includegraphics[width=0.9\linewidth]{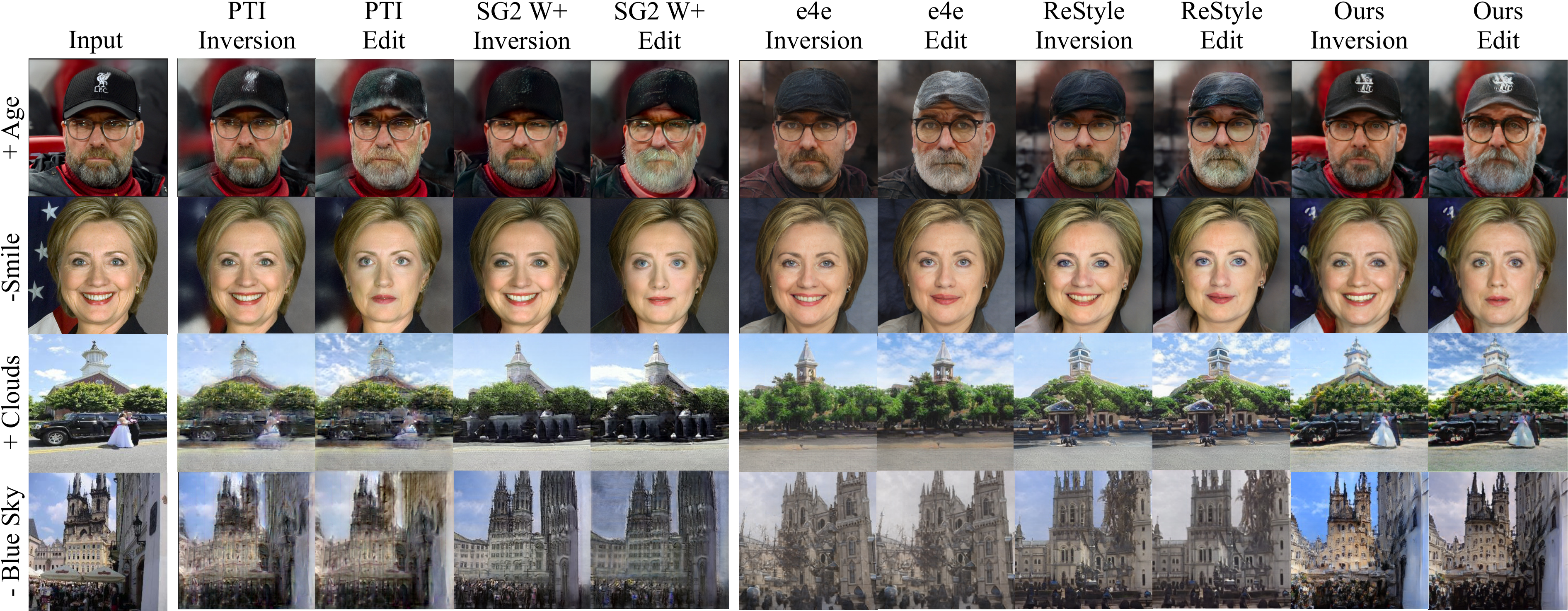}
\caption{\textbf{Qualitative editing comparison} of our method with existing StyleGAN inversion works. Editing directions on human facial domain are obtained from InterFaceGAN~\cite{shen2020interpreting}, while those for church domain are from GANSpace~\cite{harkonen2020ganspace}. More visual examples can be found in supplementary material.}
\label{fig:editing_comparison}
\vspace{-2mm}
\end{figure*}

\noindent\textbf{Qualitative Results.} 
We show the qualitative results for editing in Figure~\ref{fig:editing_comparison}. 
As our reconstruction is generally more robust than other encoder-based methods (e4e, ReStyle), it also allows better editing results. As can be seen, our work can perform reasonable edits while still preserving faithfully non-editing attributes, e.g., background. In comparison to SG2 $\mathcal{W}^{+}$, our method produces significant editing effects with fewer artifacts, e.g., beard. The reason is that our work uses the $\mathcal{W}$ space, which has excellent editing ability, while SG2 $\mathcal{W}^{+}$ uses the extension $\mathcal{W}^{+}$ space, which suffers from a lower editability. Since PTI uses latent code on $\mathcal{W}$ space to edit, so as same as our method, on the human facial domain, they can do editing operations quite well. However, in the church domain, as PTI suffers from the distortion-perception trade-off, the reconstructed images are not realistic, leading to later editing of those images is not impressive.

\subsection{User survey} 
We further conduct a user survey to evaluate the reconstruction and editing ability of our method through human perceptual assessment. In this survey, we only compare our method with other state-of-the-art encoder-based inversion ones including e4e~\cite{tov2021designing}, and ReStyle~\cite{alaluf2021restyle}. We skip comparing with pSp~\cite{richardson2021encoding} since Tov et al.~\cite{tov2021designing} pointed out that the latent code inverted by pSp is not good for editing, and they proposed e4e to fix this weakness. The setup of this user survey can be found in the supplementary. Figure~\ref{fig:user_study} demonstrated our human evaluation results. As can be seen, our method outperforms e4e and ReStyle by a large margin on both reconstruction and editing quality. Notably, our method gets favored in $71.8\%$ of reconstruction tests, $2.5$ times e4e and ReStyle combined.

\begin{figure}[h]
\centering
\includegraphics[width=\linewidth]{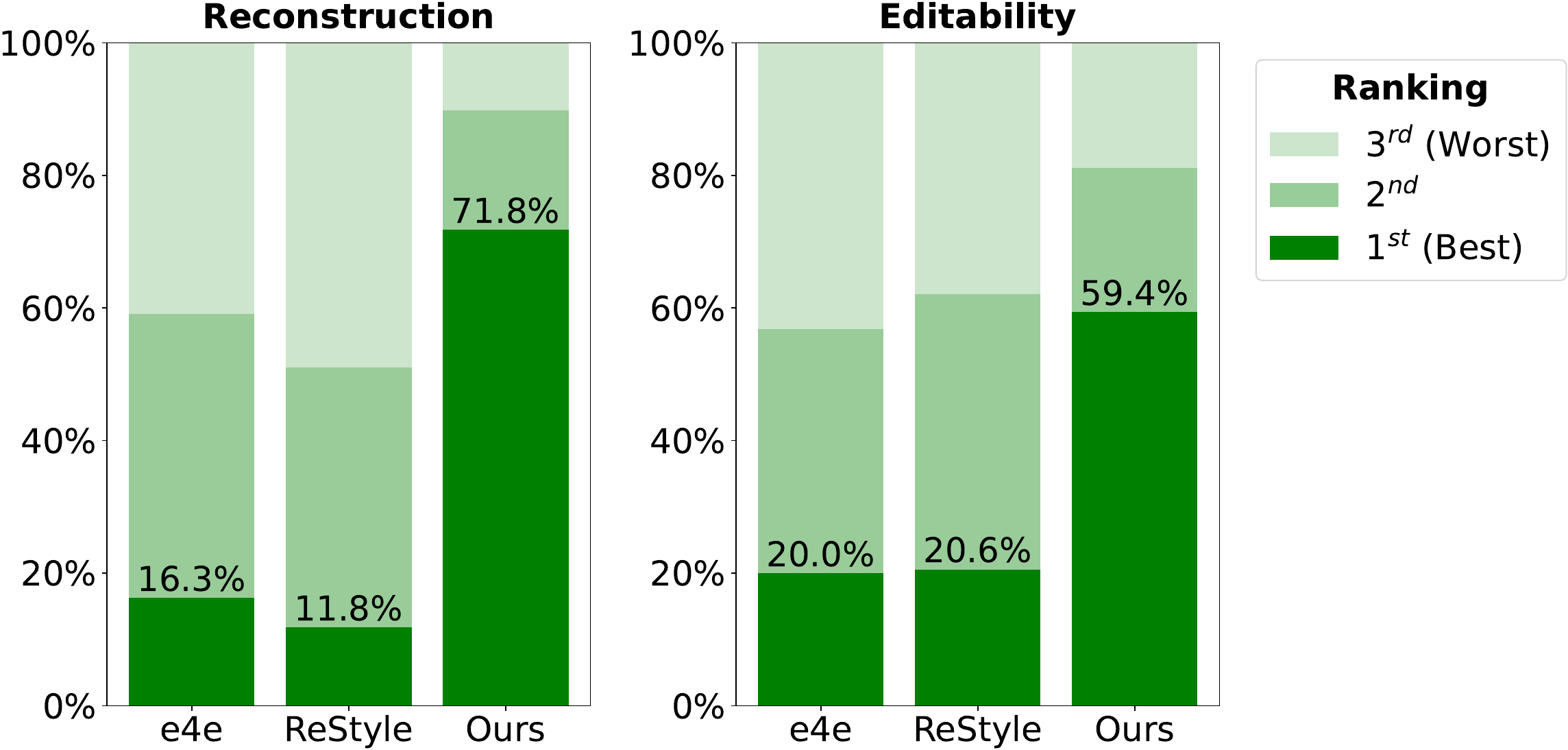}
\caption{\textbf{User study results.} We reported the percentage of times testers rank the method at $1^{st}$ (best), $2^{nd}$, and $3^{rd}$ (worst) based on two criteria, which are reconstruction and editing quality. As can be seen, our method outperforms e4e and ReStyle with a large gap.}
\label{fig:user_study}
\end{figure}

\subsection{Ablation Study}

\begin{table}
\small 
\begin{center}
\resizebox{\columnwidth}{!}{%
\begin{tabular}{l l c c c c} 
\toprule
No. & Method & L2 ($\downarrow$)  & LPIPS ($\downarrow$) & ID ($\uparrow$) & PSNR ($\uparrow$)  \\
\midrule
(1) & Ours (full) & \textbf{0.0243} & \textbf{0.1054} & \textbf{0.6013} & \textbf{21.4417} \\
(2) & (1) w/o Phase II  & 0.0539 & 0.2145 & 0.3983 & 18.6522  \\
(3) & (1) w/o $\hat{x}_{w}$ feat. & 0.0313 & 0.1264 & 0.5412 & 20.4963  \\
(4) & (1) w/o layer-wise & 0.0287 & 0.1173 & 0.5768 & 20.7693 \\
\midrule
(5) & $D=32$ & 0.0378 & 0.1492 & 0.5251 & 19.8097 \\
(6) & $D=64$ & 0.0308 & 0.1285 & 0.5571 & 20.5361 \\
(7) & $D=128$ & 0.0253 & 0.1103 & 0.5939 & 21.2115 \\
(8) & $D=256$ & \textbf{0.0243} & \textbf{0.1054} & \textbf{0.6013} & \textbf{21.4417} \\
\bottomrule
\end{tabular}
}
\end{center}
\vspace{-0.2in}
\caption{\textbf{Ablation study} on human faces.}
\label{tab:abalation_study}
\vspace{-4mm}
\end{table}

\begin{figure}[t]
\centering
\includegraphics[width=0.9\linewidth]{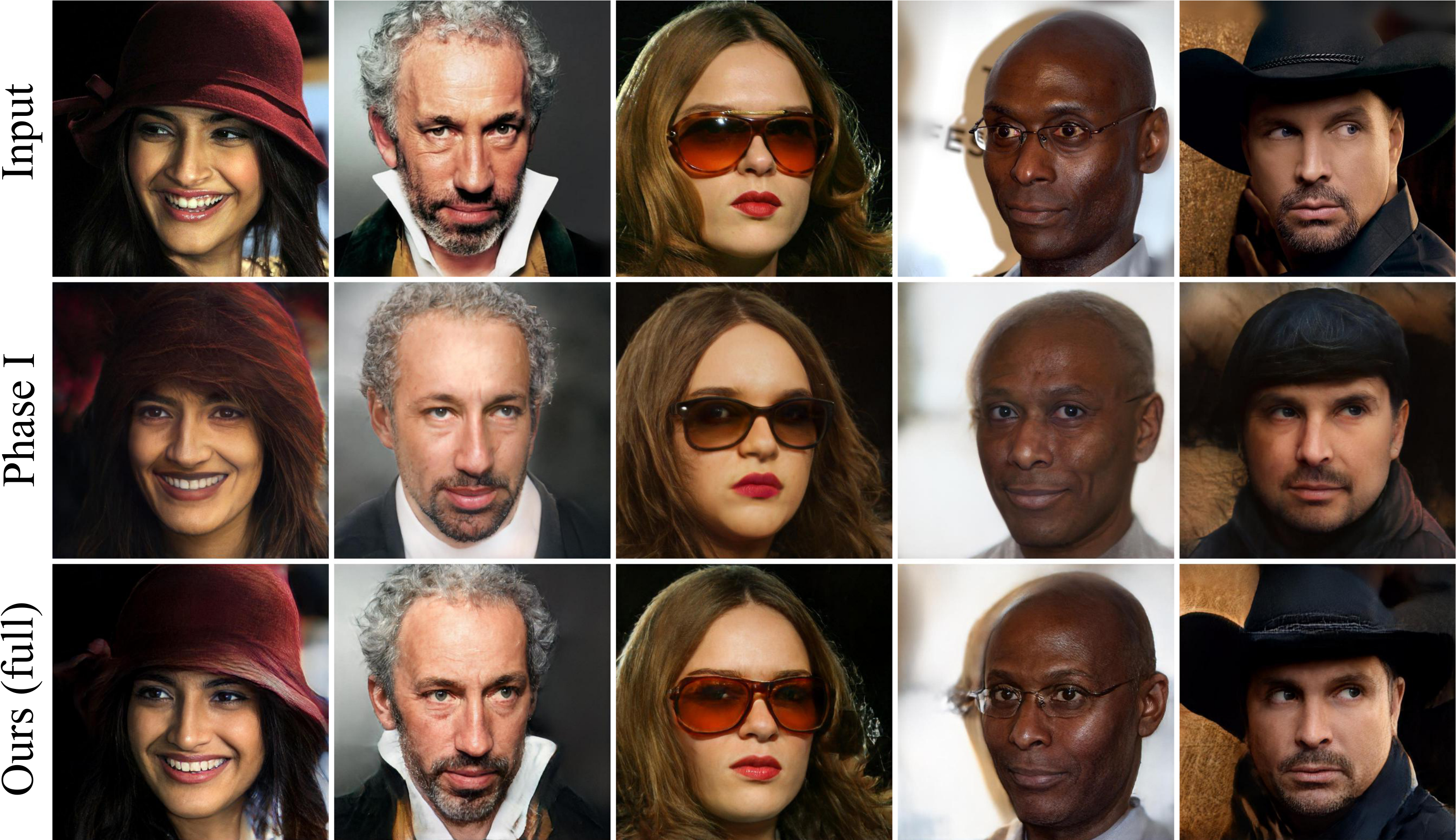}
\vspace{-1mm}
\caption{Visual examples demonstrating the effectiveness of our proposed second phase. Best viewed in zoom.}
\label{fig:ablation_study_face}
\vspace{-4mm}
\end{figure}

\noindent\textbf{Is Phase II required?} Figure~\ref{fig:ablation_study_face} and the row (2) in Table~\ref{tab:abalation_study} demonstrate the importance of our phase II in improving the reconstruction quality. As can be seen, without this phase, the reconstruction quality is significantly dropped, illustrated by all quantitative metrics. In Figure~\ref{fig:ablation_study_face}, we can see that the reconstructed images lose various details such as hat, shadowing, background, makeup, 
and more. In contrast, with the refinement from phase II, our method can recover the input image faithfully. 

\noindent\textbf{Should we fuse the appearance codes?} We test the effectiveness of our design for extracting and using the appearance code in Phase II. Firstly, we compare our full version, which uses the appearance code $h$ in the layer-wise design ($h \in \mathbb{R}^{L \times 1024 \times 8 \times 8}$), with the code-sharing version ($h \in \mathbb{R}^{ 1024 \times 8 \times 8}$).
Secondly, we test the effectiveness of using both the features from the input image $x$ and the initial image $\hat{x}_{w}$ in computing appearance code $h$. We compare our method with the version using only the features from the input image $x$. 
The results of these experiments are shown in rows (3) and (4) in Table~\ref{tab:abalation_study}, respectively. As can be seen, our full version significantly surpasses the simplified ones.

\noindent\textbf{Different choices of hyperparameter $D$ in hypernetwork.} We empirically find that the hidden dimension $D$ in our hypernetworks's design has a remarkable effect on the model performance. Rows (5), (6), (7), (8) in Table~\ref{tab:abalation_study} investigates different choices of $D$, proving that $D=256$ is best option.

\begin{figure}[h]
\centering
\includegraphics[width=\linewidth]{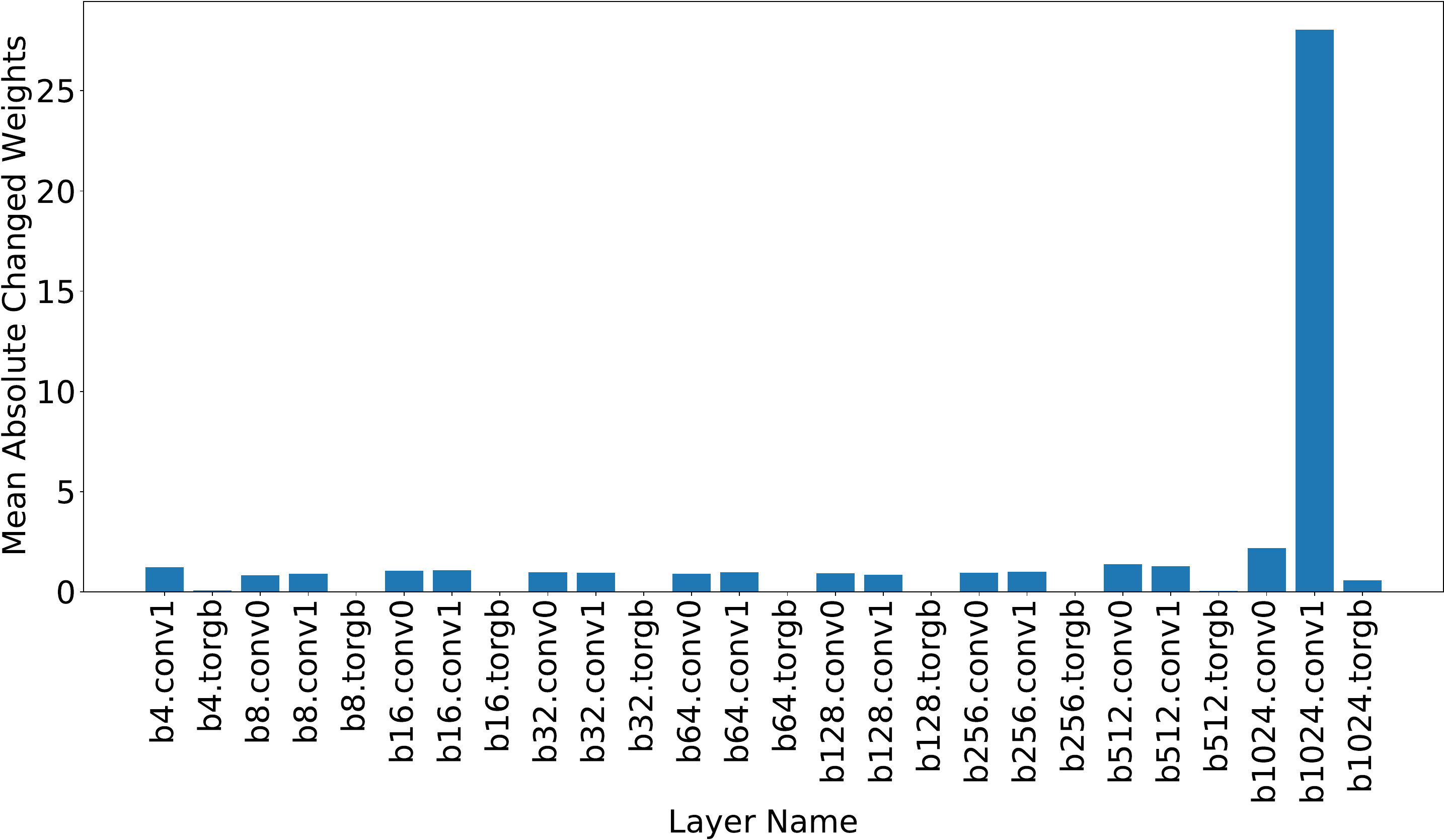}
\caption{Visualizing the statistic of residual weights.}
\label{fig:weight_change_statistic}
\vspace{-2mm}
\end{figure}

\noindent\textbf{Which layers in the generator are updated?} We also visualize the residual weights predicted by the hypernetworks to analyze which StyleGAN layers change the most when applying Phase II. Specifically, we choose to visualize the mean absolute amount of weight change between the parameters of each layer and compare this value with other layers. This statistic is calculated from the residual weights predicted by our method on $2,824$ images in the CelebA-HQ test set. Figure~\ref{fig:weight_change_statistic} visualizes this analysis. From this figure, we draw two following insights. First, the convolution layers in main generator blocks contribute more significantly than the convolution layers in \textit{torgb} blocks. Second, the weight change for the last resolution ($1024 \times 1024$) is significantly larger than the ones for other resolutions. As we expected, on human faces, since the first phase reconstructed images quite well overall, therefore in phase II, they should mainly focus on refining the fine-grained details. This argument is supported well by insight 2.

\subsection{Application: Real-world Image Interpolation}

\begin{figure}[t] 
\centering
\vspace{-1mm}
\includegraphics[width=\linewidth]{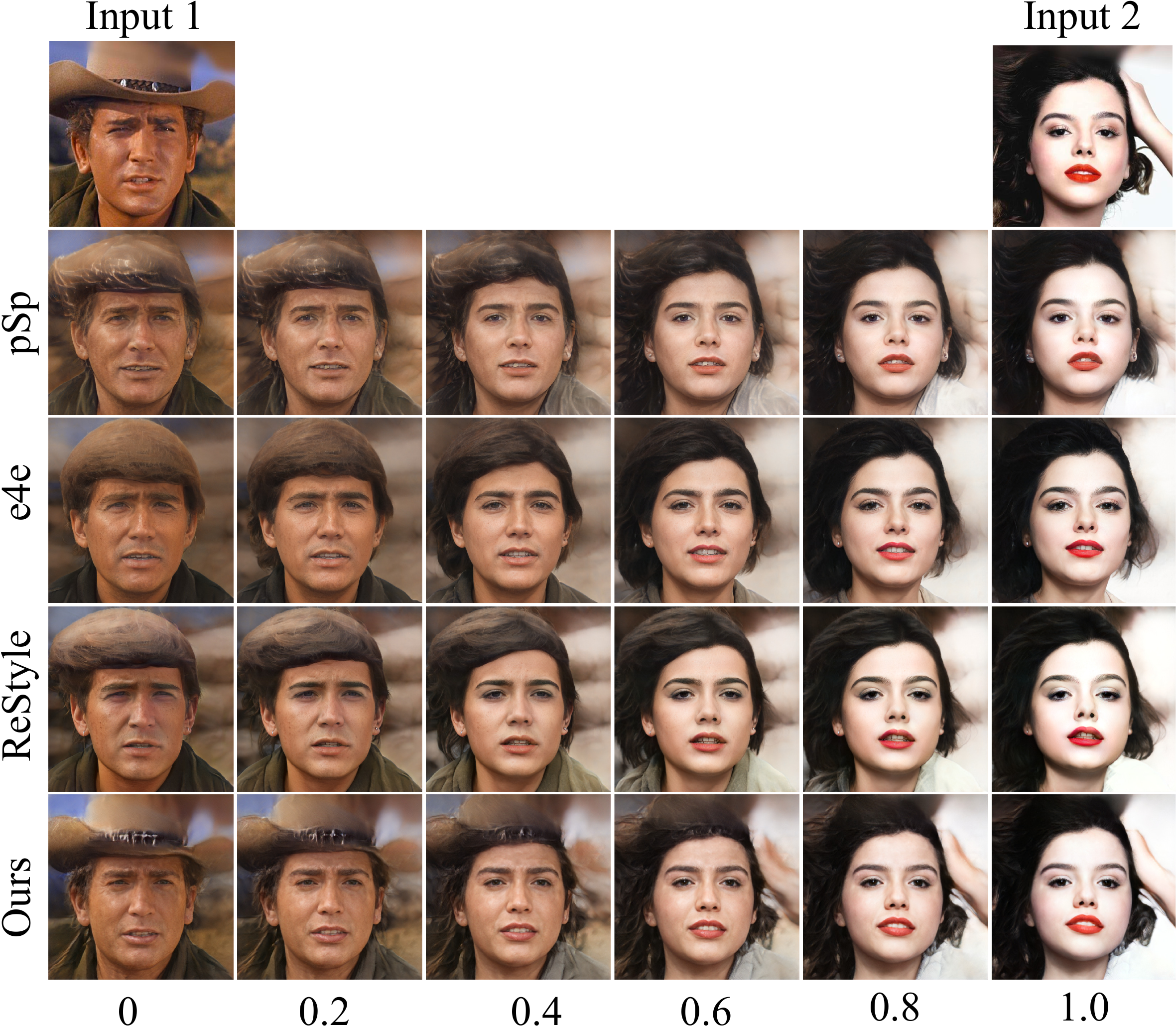}
\caption{Real-world image interpolation results.}
\label{fig:interpolation}
\vspace{-5mm}
\end{figure}

In this section, we present a new approach to interpolate between two real-world images. Given two input images, a common-used process is first to find the corresponding latent codes via GAN inversion, compute the linear interpolated latent codes, and pass it through the GAN model to get the interpolated images. Motivated by our two-phase inversion mechanism, we propose a new interpolation approach by \emph{interpolating both latent codes and generator weights} instead of using only latent codes as the common existing pipeline. Figure~\ref{fig:interpolation} shows some qualitative results comparing our method with the ones interpolating only latent codes. As can be seen, our method not only reconstructed input images with correctly fine-grained details (e.g., hat, background, and hand) but also provided the smooth interpolated images.

\section{Conclusion}
This paper presented a fully encoder-based method for solving the StyleGAN inversion problem using a two-phase approach and the hypernetworks to refine the generator's weights. As a result, our method has high fidelity reconstruction, excellent editability while running almost real-time.

Despite the promising results, our work is not without limitations. As an encoder-based method, it generates high-quality images but appears not completely better than approaches based on per-image optimization such as SG2 $\mathcal{W}^{+}$, PTI in terms of metrics. However, we also include some cases in Figure~\ref{fig:reconstruction_comparison} to show that both SG2 $\mathcal{W}^{+}$ and PTI can yield inaccurate reconstruction. We advocate further development of encoder-based methods to reduce such a performance gap since encoder-based methods (including ours) run faster at inference, which makes them more suitable for interactive and video applications. Specially, our work can be further improved by applying iterative refinement ~\cite{alaluf2021restyle}, multi-layer identity loss~\cite{wei2021simpleinversion}, or extending to StyleGAN3~\cite{karras2021alias}.

{
\small
\bibliographystyle{plainnat}
\bibliography{egbib}
}

\appendix
\appendixpage

In this document, we first discuss the potential negative social impacts of our research. Then, we present another user study for the churches domain. This survey and the previous one presented in the main manuscript for human faces show that our method works well for both domains under the human assessment. Moreover, we present the difference map visualizations to analyze the image residuals from our Phase II update. We also include additional dataset and implementation details for reproducibility. Finally, we provide extensive visual examples for further qualitative evaluation. 

\section{Discussion on Negative Societal Impacts}
Besides some fancy and potential applications that could be commercial in the future to create numerous profits and have a large impact on society, our method can not avoid being used in ways harmful to society in some cases. For example, an attacker can use our model to create deep fake examples from reconstructing and editing a real photograph of a human face. Our inversion method can potentially yield realistic results, making them indistinguishable from real faces. Our method might also lead to the caveat of creating deep fake videos at real time since our neural network performs very fast predictions. 
Despite such, we believe that our method could contribute positively to the society via inspiring and advocating the development of more sophisticated detection methods to mitigate deep fakes.

\section{User Study} 
\subsection{The Churches Domain}
\begin{figure}[t]
\centering
\includegraphics[width=\linewidth]{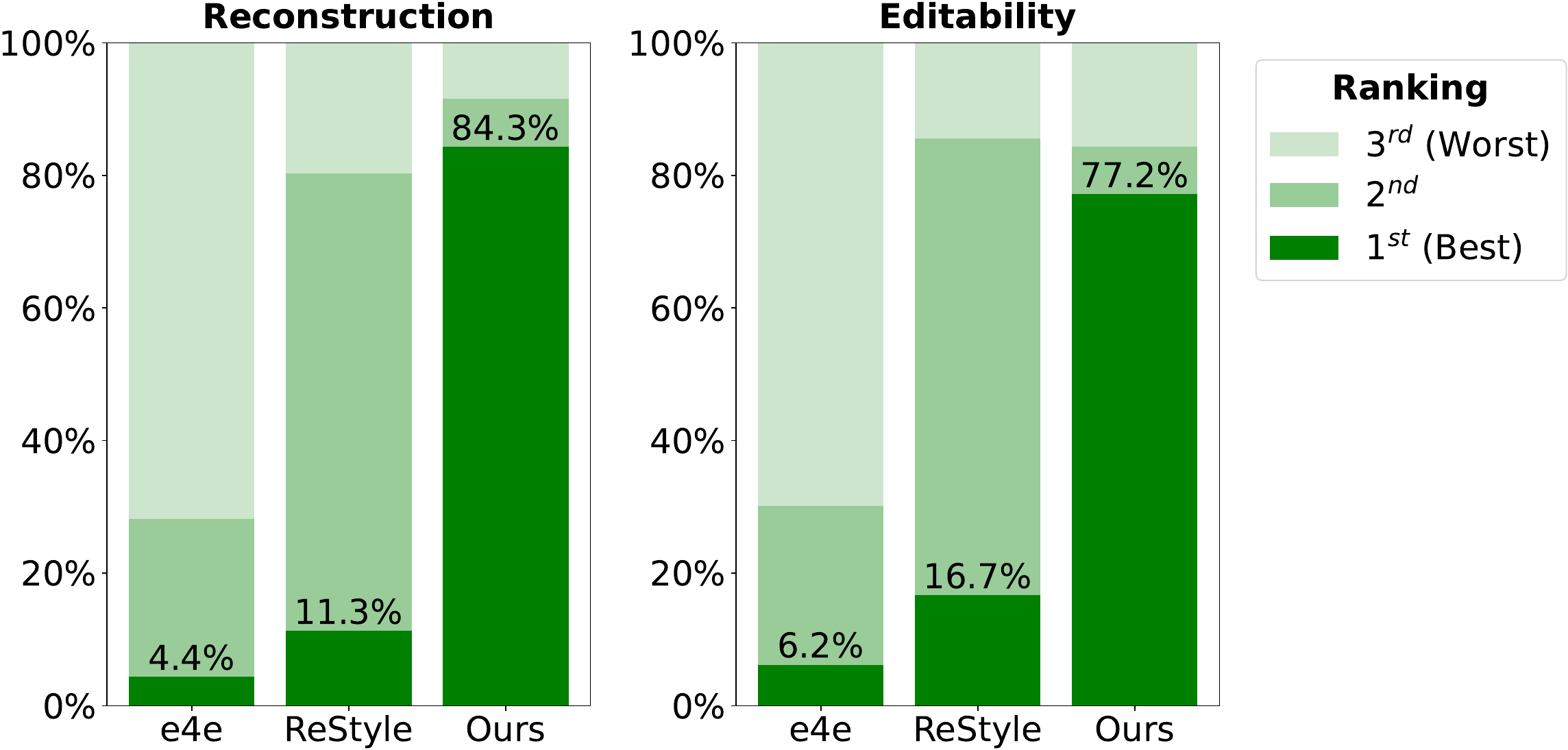}
\vspace{-1mm}
\caption{\textbf{User study results for Churches domain.} We reported the percentage of times testers rank the method at $1^{st}$ (best), $2^{nd}$, and $3^{rd}$ (worst) based on two criteria, which are reconstruction and editing quality. As can be seen, our method outperforms e4e and ReStyle significantly.}
\vspace{-4mm}
\label{fig:church_user_study}
\end{figure}

In the main paper, we have shown the user study results for the human faces domain. Here, we also conduct an additional user study on the churches domain to verify the effectiveness of our method on this domain. The results of the user study for churches domain can be found in Figure~\ref{fig:church_user_study}. As can be seen, our method outperforms significantly e4e~\cite{tov2021designing} and ReStyle~\cite{alaluf2021restyle} with a very large gap. Our method is favored in $84.3\%$ and $77.2\%$ of the reconstruction and editing tests, respectively, which is about $5.3\times$ and $3.3\times$ better than e4e and ReStyle combined for each test.

\subsection{Details of User Study}
We now turn to describe the detailed setup of our user survey. Particularly, there are two separate tests, which are \emph{reconstruction} and \emph{editability} tests. For each test, we ask each human subject to rank the image from the scale of $1^{st}$ (best) to $3^{rd}$ (worst) based on some criteria, which depend on the corresponding domain and would be described below.

\noindent \textbf{Human faces.} We first choose $30$ images from the images of the CelebA-HQ test set and the collected high-resolution images from the Internet as the input images. Then, we use the candidate methods to reconstruct and edit these input images. Each question is a record of input image and the reconstructed/edited images by the candidate methods. 
For reconstruction, we ask each participant to rank the images on: (1) the ability to preserve identity; (2) the ability to reconstruct the details such as background, makeup, shadow, lipstick, hat, etc; (3) image aesthetics. 
For editability, we request the participant to rank: (1) the level of identity preservation compared to the input image; (2) the ability to preserve the details (except for the editing attribute) of the original photo in the edited image -- the more details the better. We recruited a total of $38$ and $30$ participants to cast $1,140$ and $900$ votes for the reconstruction and editing tests, respectively. 

\noindent \textbf{Churches.} We choose randomly $30$ images from the test set of LSUN Church dataset as the input images. Then, we use the candidate methods to reconstruct and edit these input images. Each question is a record of the input image and the reconstructed/edited images by the candidate methods. 
For reconstruction, we ask each human subject to rank on: (1) the ability to restore as much as details of the input image in the reconstructed image; (2) image aesthetics. 
For editability, we request the participant to rank on the ability to preserve as much detail of the input image as possible (except for the editing attribute). 
We recruited $35$ and $26$ participants, which results in $1,050$ and $780$ votes for the reconstruction and editing tests, respectively. 

\section{Visualization and Analysis}
\subsection{Distribution of the predicted residual weights}
\begin{figure}[h]
\centering
\includegraphics[width=0.9\linewidth]{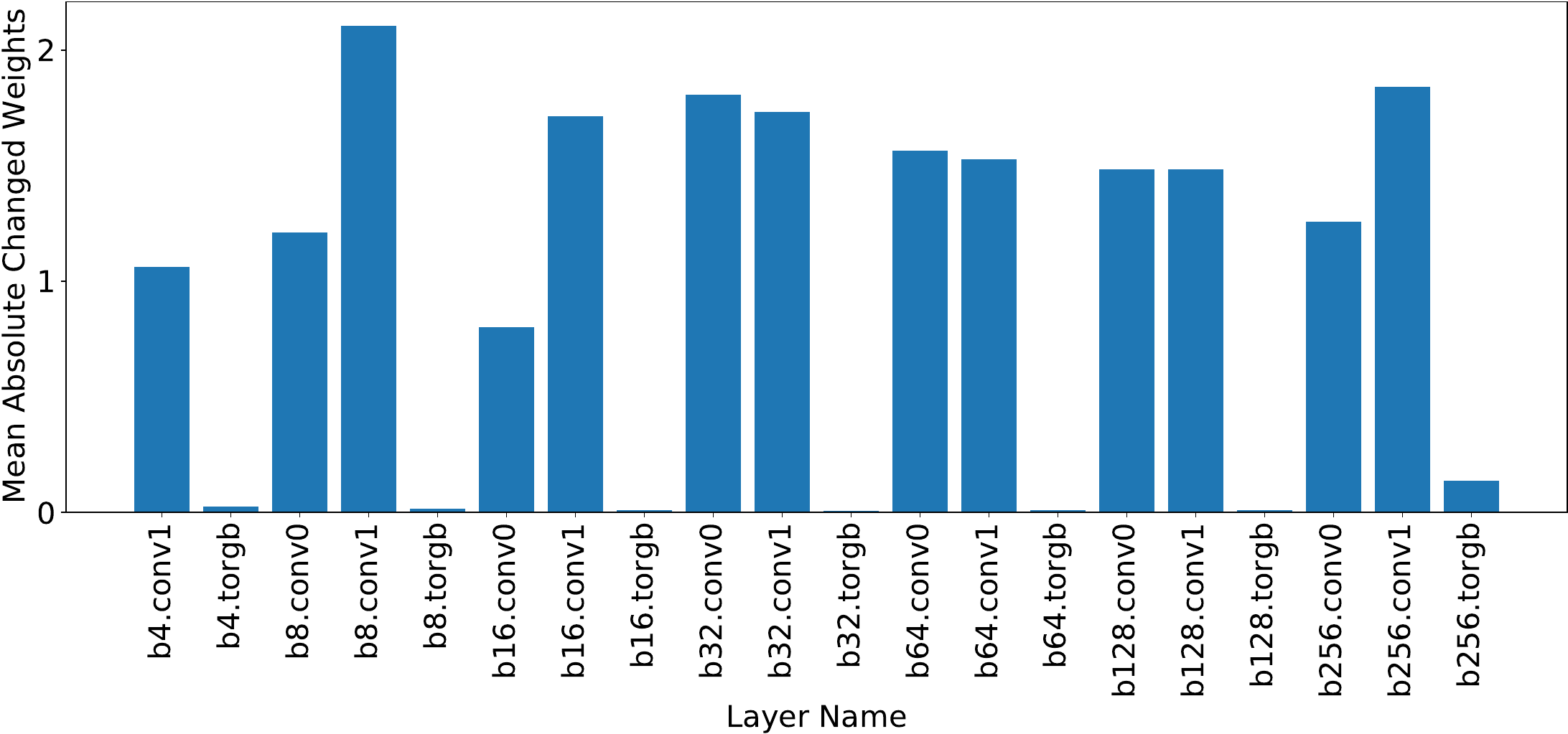}
\vspace{-2mm}
\caption{Visualizing the statistic of residual weights predicted by the hypernetworks on the churches domain. Compared to faces, our hypernetworks provides more uniform weight updates across layers, which means updates are required on both low- and high-frequency signals of the images.}
\label{fig:church_weight_change_statistic}
\end{figure}

In the main paper, we have shown an analysis on the distribution of residual weights predicted by the hypernetworks in Phase II for the human facial domain. Here, we also provide an equivalent analysis for the churches domain. Figure~\ref{fig:church_weight_change_statistic} presents this visualization. As can be seen, the churches domain results are not completely the same as those for the human facial domain. 
The observation for human faces still preserved in the churches domain is that the \emph{main conv} weights contribute significantly compared to weights of \emph{torgb} block. The difference here is that the residual weight updates occur at most layers in churches instead of concentrating at the last layer as in human faces. This aligns with the fact that the church domain is more diverse and challenging to reconstruct than the human facial domain, thus requiring updates at both low- and high-frequency signals. For face images, the initial images after Phase I are already very close to the inputs, and Phase II focuses on restoring the fined-grained details.

\subsection{Difference maps}
\begin{figure}[t]
    \captionsetup[subfigure]{justification=centering}
    \begin{subfigure}[b]{.49\columnwidth}
    \centering
    \includegraphics[width=0.9\linewidth]{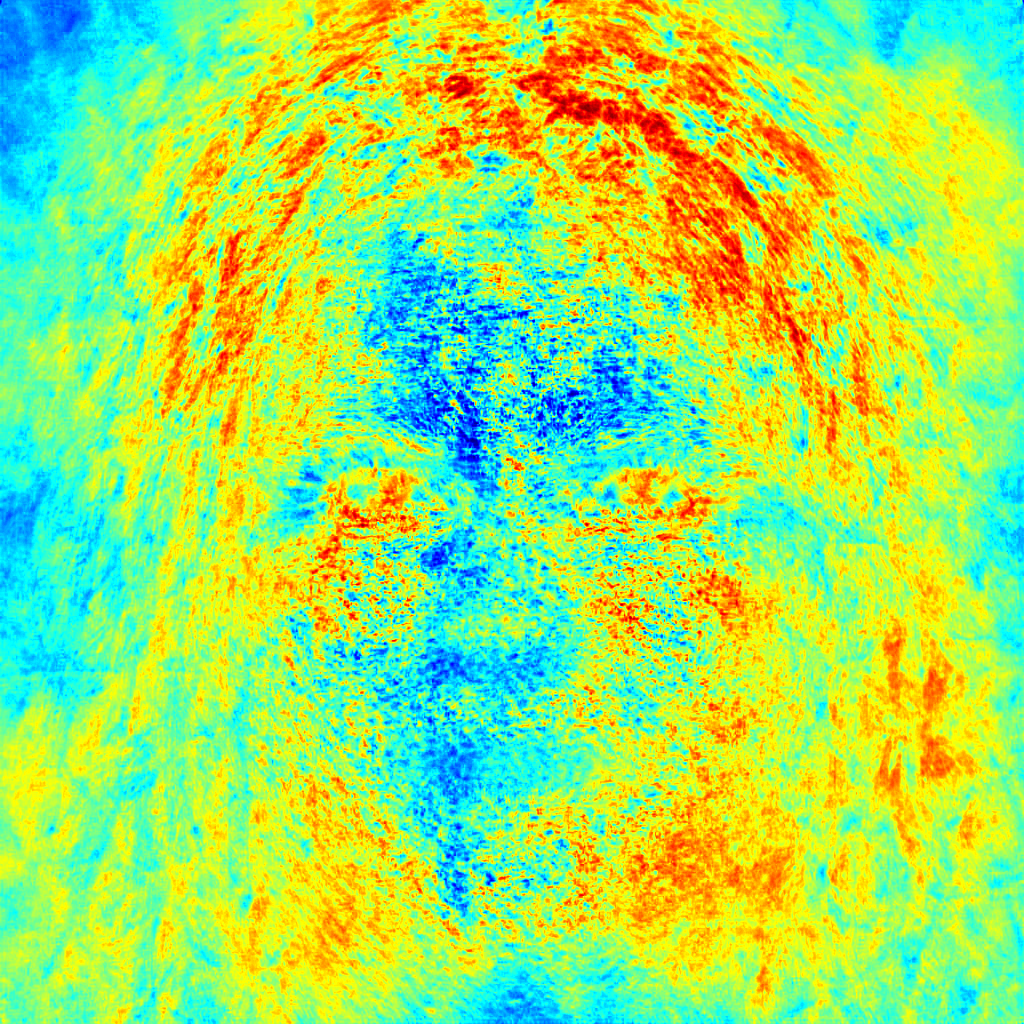}
    \caption{\textbf{Human Faces}}
    \end{subfigure}
    \begin{subfigure}[b]{.49\columnwidth}
    \centering
    \includegraphics[width=0.9\linewidth]{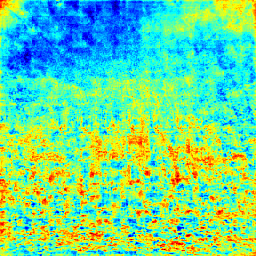}
    \caption{\textbf{Churches}}
    \end{subfigure}
    \caption{\textbf{Heat-map visualization} of the difference between the output images in Phase I and the final images after both phases of our method, averaged over images from the test set. For faces, large changes are focused around eye and hair. For churches, low changes are in the sky region. 
    \textcolor{blue}{Blue} indicates a small change, whereas \textcolor{red}{red} denotes a large change.}
    \label{fig:heat_map}
\end{figure}

We also conduct an experiment to visualize the difference between the initial image from Phase I and the final image from both phases to analysis which regions change most in the refinement process of Phase II. Specifically, given the input image $x^{(i)}$, the Phase I's output image $\hat{x}_{w}^{(i)}$, and the final reconstructed image $\hat{x}^{(i)}$, where $i \in \{1..N\}$, $N$ is the number of test images, we compute the difference map $m^{(i)}$ by subtracting two reconstructed images and take absolute values, which means $m^{(i)} = |\hat{x}^{(i)} - \hat{x}_{w}^{(i)}|$. We then compute the mean difference map $\overline{m}$ by averaging all difference maps $\{m^{(1)}, m^{(2)}, ,m^{(N)}\}$. Next, we convert $\overline{m}$ from the RGB image to the grayscale image and visualize it as the heat map. In this experiment, we use all $2,824$ images from the CelebA-HQ test set and all $300$ images from LSUN Church test set to analyze for human faces and churches domains, respectively. Figure~\ref{fig:heat_map} presents the results. As can be seen, for human faces, the heat map in Figure~\ref{fig:heat_map}-a reveals that our method focuses mainly on refining the regions having many fine-grained details such as hair, cheek, beard, eye in Phase II. For the churches domain, since the images from this domain are not aligned and very diverse in terms of structures, its heat map in Figure~\ref{fig:heat_map}-b appears more random. However, we can see that the top left region of the map does not change much, which is often the sky that is already well reconstructed from Phase I. 

\begin{figure}[h]
\centering
\includegraphics[width=\linewidth]{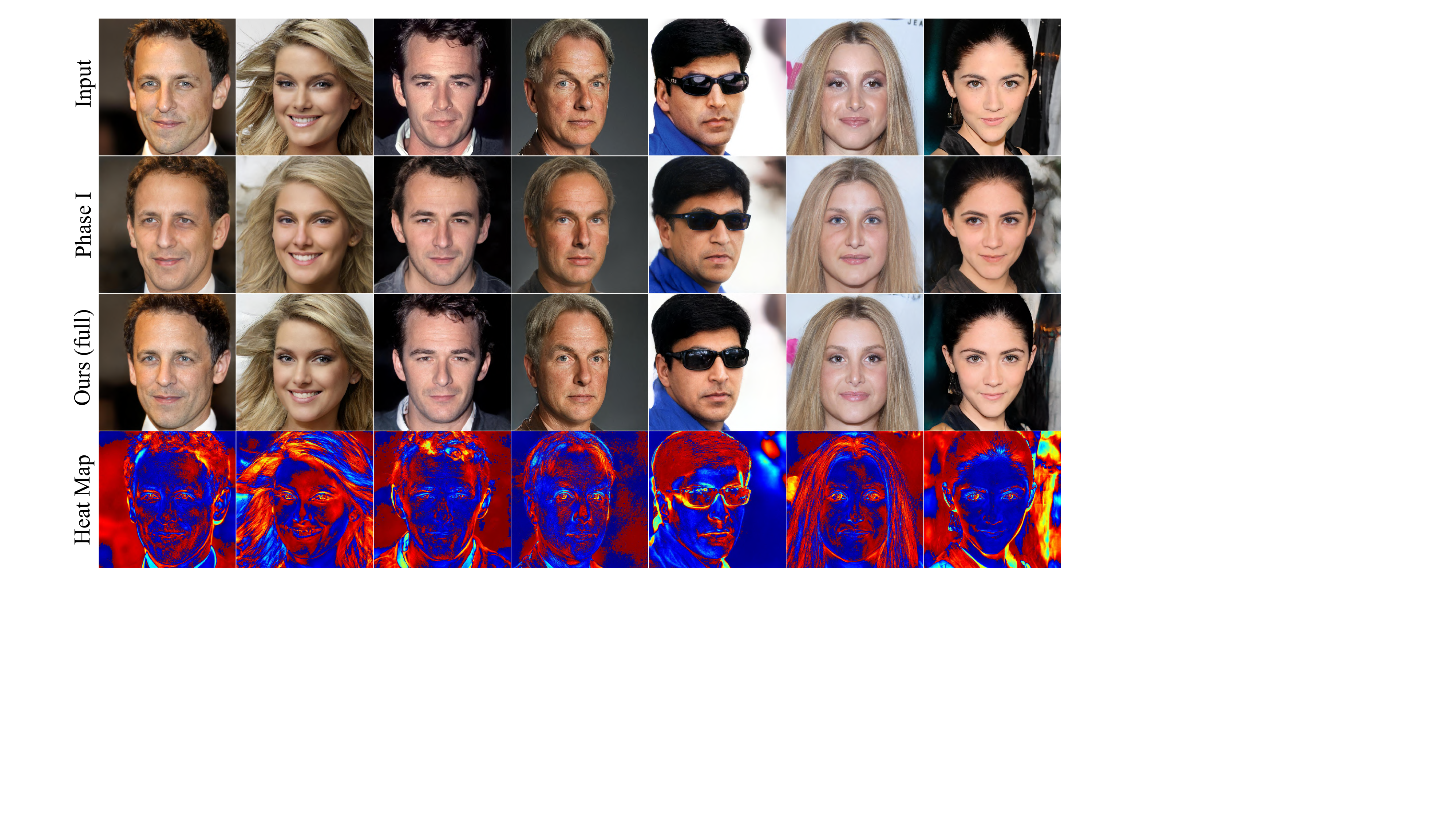}
\caption{Visualizing the effectiveness of our phase II in bringing back the information of input image missed in the initial image on the human facial domain. We also include the difference map for reference which region change most in the image after phase II. \textcolor{blue}{Blue} indicates a small change, while \textcolor{red}{red} denotes a large change. Best viewed in zoom.}
\label{fig:ablation_study_face_supp_1}
\end{figure}

\begin{figure}[h]
\centering
\includegraphics[width=\linewidth]{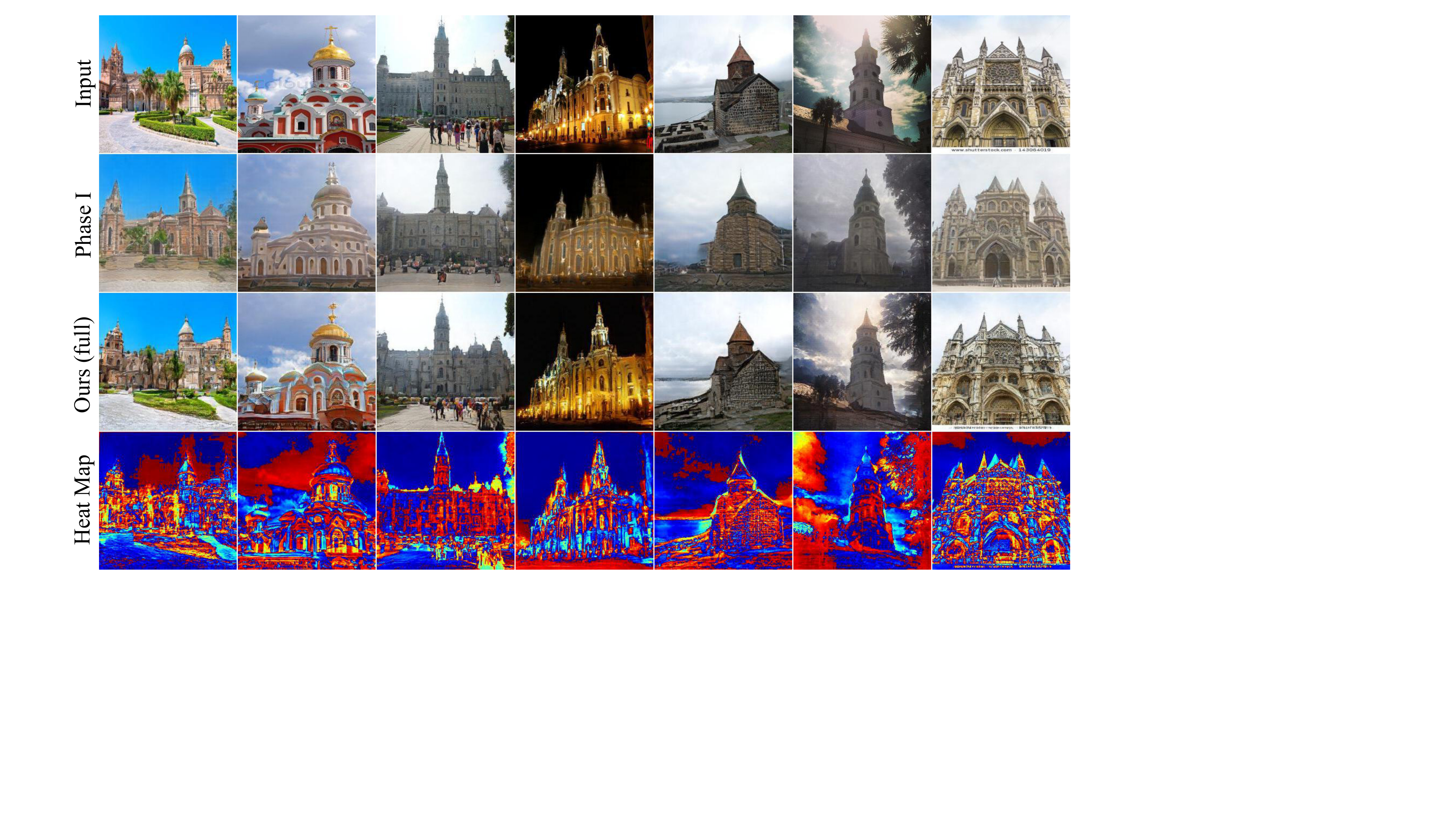}
\caption{Visualizing the effectiveness of our phase II in bringing back the information of input image missed in the initial image on the churches domain. We also include the difference map for reference which region change most in the image after phase II. \textcolor{blue}{Blue} indicates a small change, while \textcolor{red}{red} denotes a large change. Best viewed in zoom.}
\label{fig:ablation_study_church_supp_1}
\end{figure}

To gain better insights on individual cases, we also provide heat-maps for each image in Figure~\ref{fig:ablation_study_face_supp_1} and~\ref{fig:ablation_study_church_supp_1}. 

\section{Additional Experimental Details}
\subsection{Datasets}

In this section, we provide more details about the datasets we employ in conducting our experiments.

\noindent\textbf{Human faces.} We use $70,000$ images from the FFHQ~\cite{karras2019stylegan} dataset as our training set and $2,824$ images from the official test set of CelebA-HQ~\cite{liu2015deep, karras2018progressive} as our test set. These datasets contain high-quality real-world face images at resolution 1024x1024. All faces are aligned to the center of the images.

\noindent\textbf{Churches.} We choose the Churches domain to test our method on images of natural outdoor scenes. These images are more diverse than human faces and thus considered more challenging. We use LSUN Church~\cite{yu2015lsun} in this task. The resolution of images is $256 \times 256$. We use $126,227$ and $300$ images from the official train/test split of LSUN Church for training and testing, respectively. 

\subsection{Detailed architecture of $E_1$ and $E_2$ encoders.}
\begin{figure}[h]
\centering
\includegraphics[width=\linewidth]{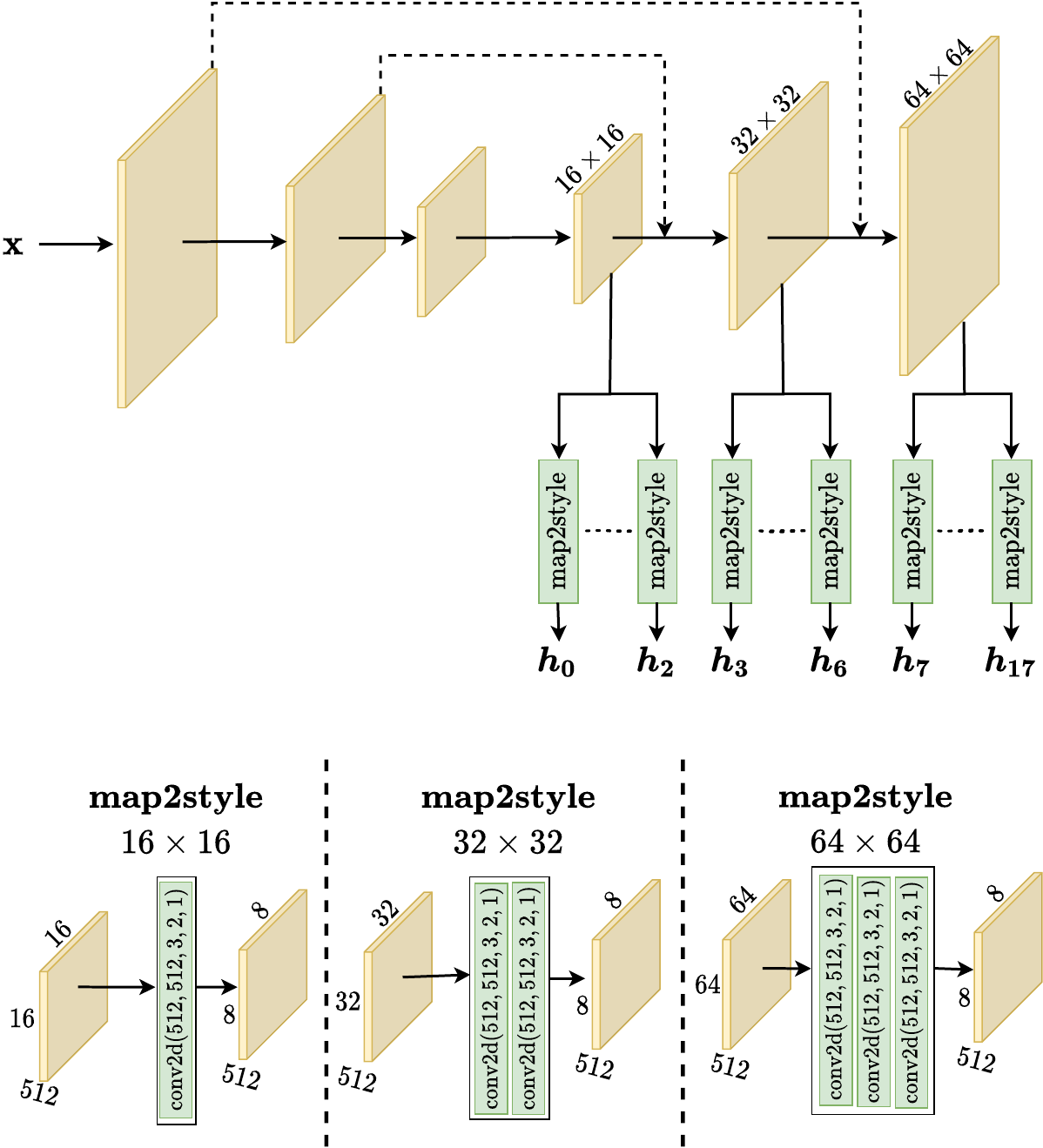}
\caption{The FPN-based encoder network proposed by Richardson et al.~\cite{richardson2021encoding}, which has been used popularly by many previous encoder-based GAN inversion works, and our modifications in \emph{map2style} block to output $512 \times 8 \times 8$ tensor instead of $512$ vector as the original backbone. Diagram notation: conv2d(in\_channels, out\_channels, kernel\_size, stride, padding). For simplicity, we omit the LeakyReLU activations after each conv2d layer in the figure. }
\label{fig:encoder_arch_supp}
\end{figure}

 As mentioned previously in the main paper, for $E_1$ and $E_2$ encoders, we adopt the design of ~\cite{richardson2021encoding, tov2021designing} as the main backbone with some modifications since the superior performance of the original network. For $E_1$ encoder, we utilize the $\mathcal{W}$ encoder of these networks without modifications. For $E_2$ encoder, since our network outputs the intermediate features having the similar size with the output of the $\mathcal{W}^{+}$ encoder of ~\cite{richardson2021encoding, tov2021designing}. Therefore we also leverage this design for $E_2$ with some modifications. Particularly, the architecture of this encoder is the FPN-based design~\cite{richardson2021encoding}. We modify the \emph{map2style} block to output the feature tensor with the dimension of $512 \times 8 \times 8$ instead of a $512$ vector of original backbone. Figure~\ref{fig:encoder_arch_supp} gives the network design of FPN-based network and our modifications. 

\section{Additional Qualitative Results}
\begin{figure*}[h]
\centering
\includegraphics[width=\linewidth]{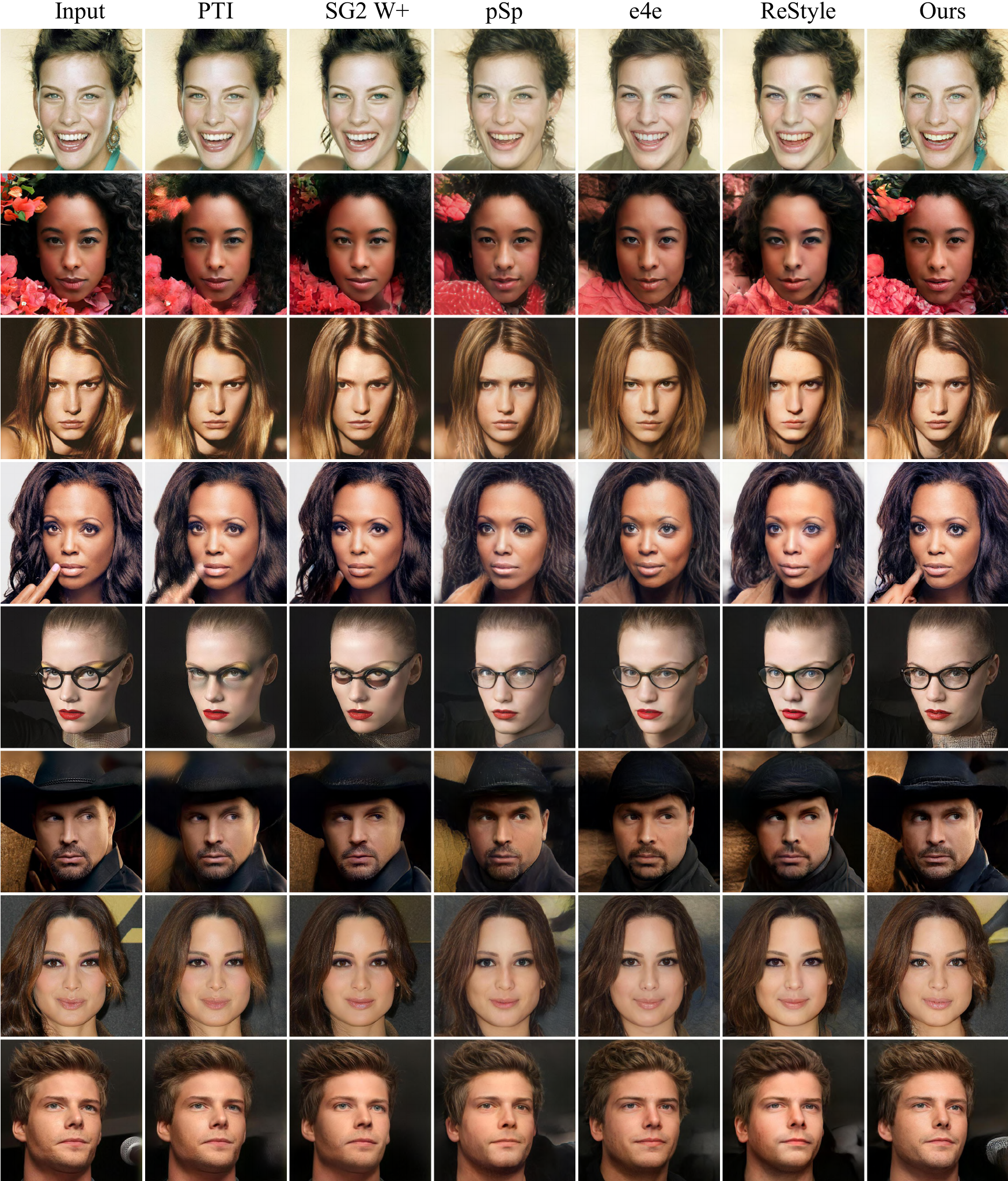}
\caption{More visual examples shown the \textbf{reconstruction comparison} of our method with other \emph{encodered-based approaches}: pSp~\cite{richardson2021encoding}, e4e~\cite{tov2021designing}, ReStyle~\cite{alaluf2021restyle}; \emph{optimization-based methods}: SG2~$\mathcal{W}^{+}$~\cite{abdal2019image2stylegan}; \emph{two-stage works}: PTI~\cite{roich2021pivotal} on the human facial domain. The input images are taken from the CelebA-HQ~\cite{liu2015deep, karras2018progressive} test set. Best viewed in zoom.}
\label{fig:face_reconstruction_supp_1}
\end{figure*}

\begin{figure*}[h]
\centering
\includegraphics[width=\linewidth]{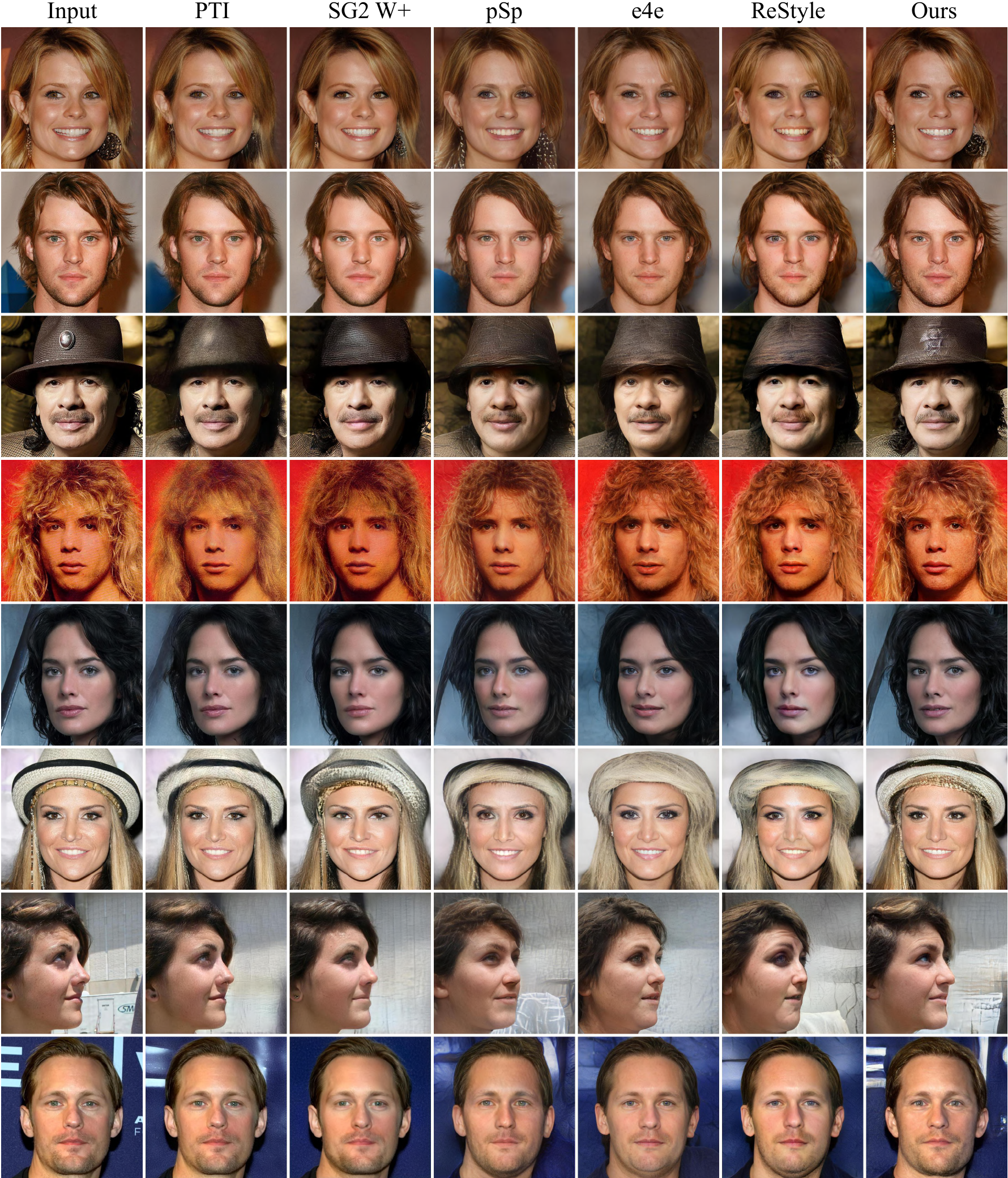}
\caption{More visual examples shown the \textbf{reconstruction comparison} of our method with other \emph{encodered-based approaches}: pSp~\cite{richardson2021encoding}, e4e~\cite{tov2021designing}, ReStyle~\cite{alaluf2021restyle}; \emph{optimization-based methods}: SG2~$\mathcal{W}^{+}$~\cite{abdal2019image2stylegan}; \emph{two-stage works}: PTI~\cite{roich2021pivotal} on the human facial domain. The input images are taken from the CelebA-HQ~\cite{liu2015deep, karras2018progressive} test set. Best viewed in zoom.}
\label{fig:face_reconstruction_supp_2}
\end{figure*}

\begin{figure*}[h]
\centering
\includegraphics[width=\linewidth]{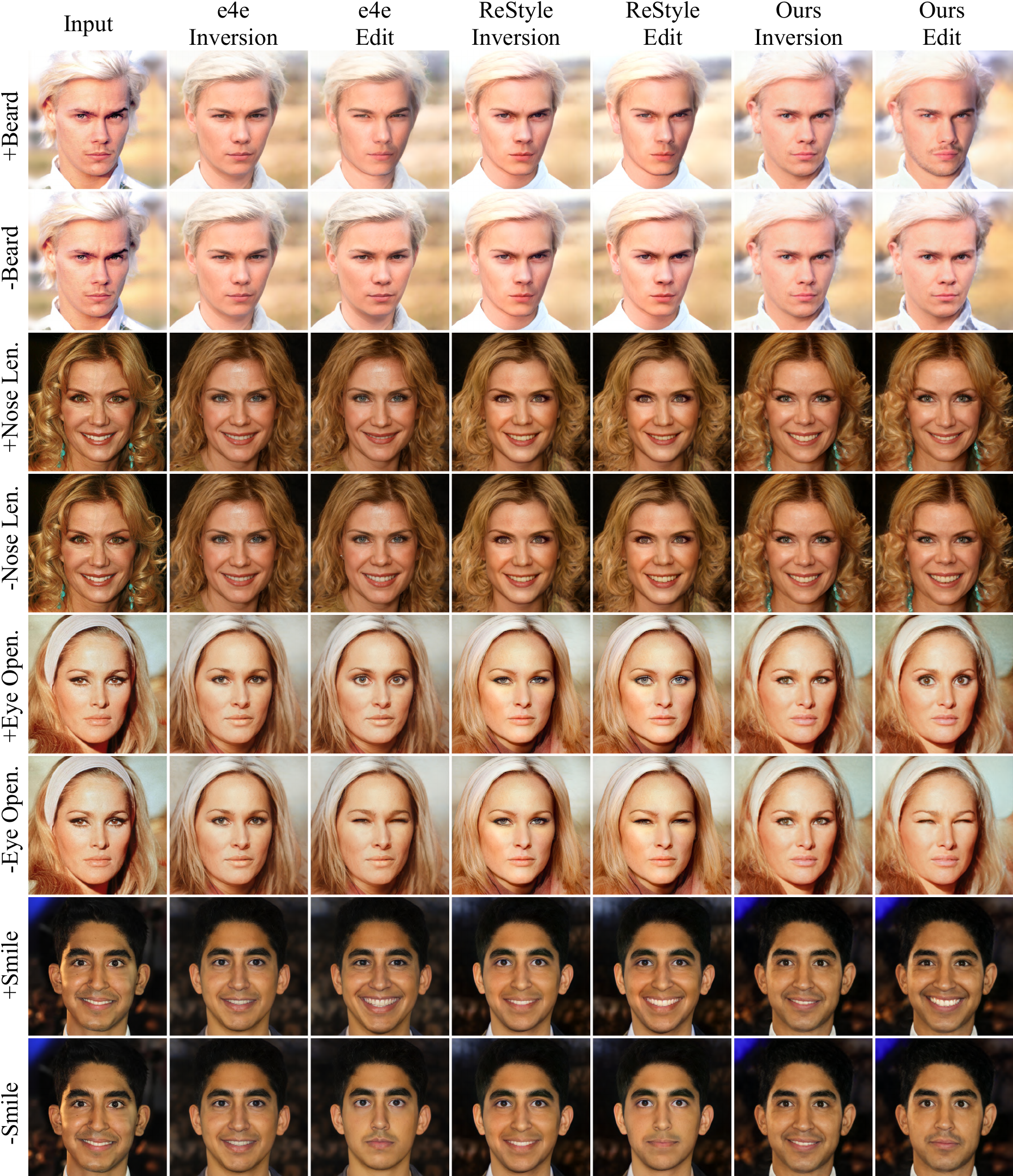}
\caption{More visual examples shown the \textbf{editability comparison} of our method with the existing \emph{encodered-based approaches}, which are e4e~\cite{tov2021designing}, ReStyle~\cite{alaluf2021restyle} on the human facial domain. The input images are taken from the CelebA-HQ~\cite{liu2015deep, karras2018progressive} test set. The smile direction is obtained from InterFaceGAN~\cite{shen2020interpreting}, whereas other directions are borrowed from GANSpace~\cite{harkonen2020ganspace}. Best viewed in zoom.}
\label{fig:encoder_based_face_edit_supp_1}
\end{figure*}

\begin{figure*}[h]
\centering
\includegraphics[width=\linewidth]{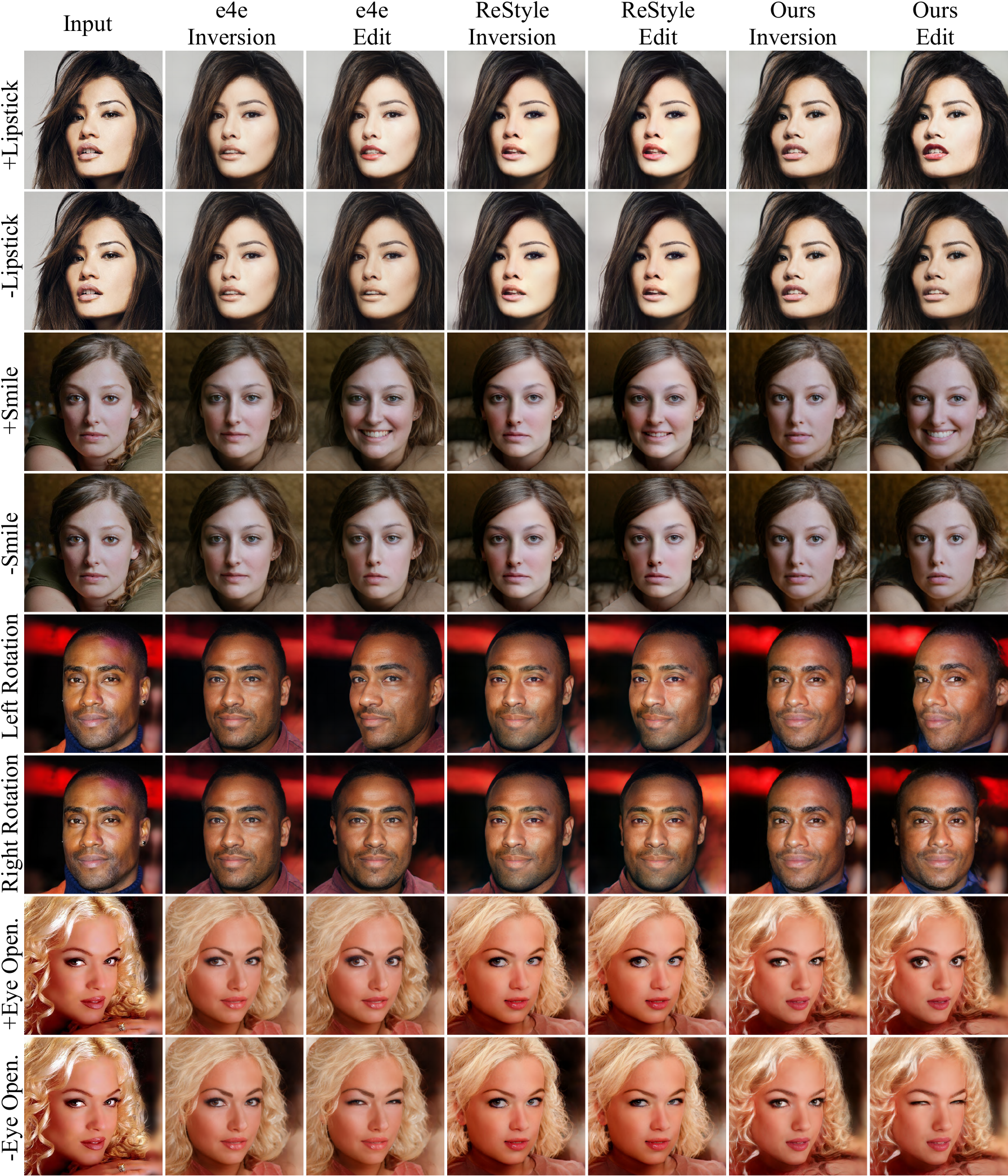}
\caption{More visual examples shown the \textbf{editability comparison} of our method with the existing \emph{encodered-based approaches}, which are e4e~\cite{tov2021designing}, ReStyle~\cite{alaluf2021restyle} on the human facial domain. The input images are taken from the CelebA-HQ~\cite{liu2015deep, karras2018progressive} test set. The rotation and smile directions are obtained from InterFaceGAN~\cite{shen2020interpreting}, whereas other directions are borrowed from GANSpace~\cite{harkonen2020ganspace}. Best viewed in zoom.}
\label{fig:encoder_based_face_edit_supp_2}
\end{figure*}

\begin{figure*}[h]
\centering
\includegraphics[width=\linewidth]{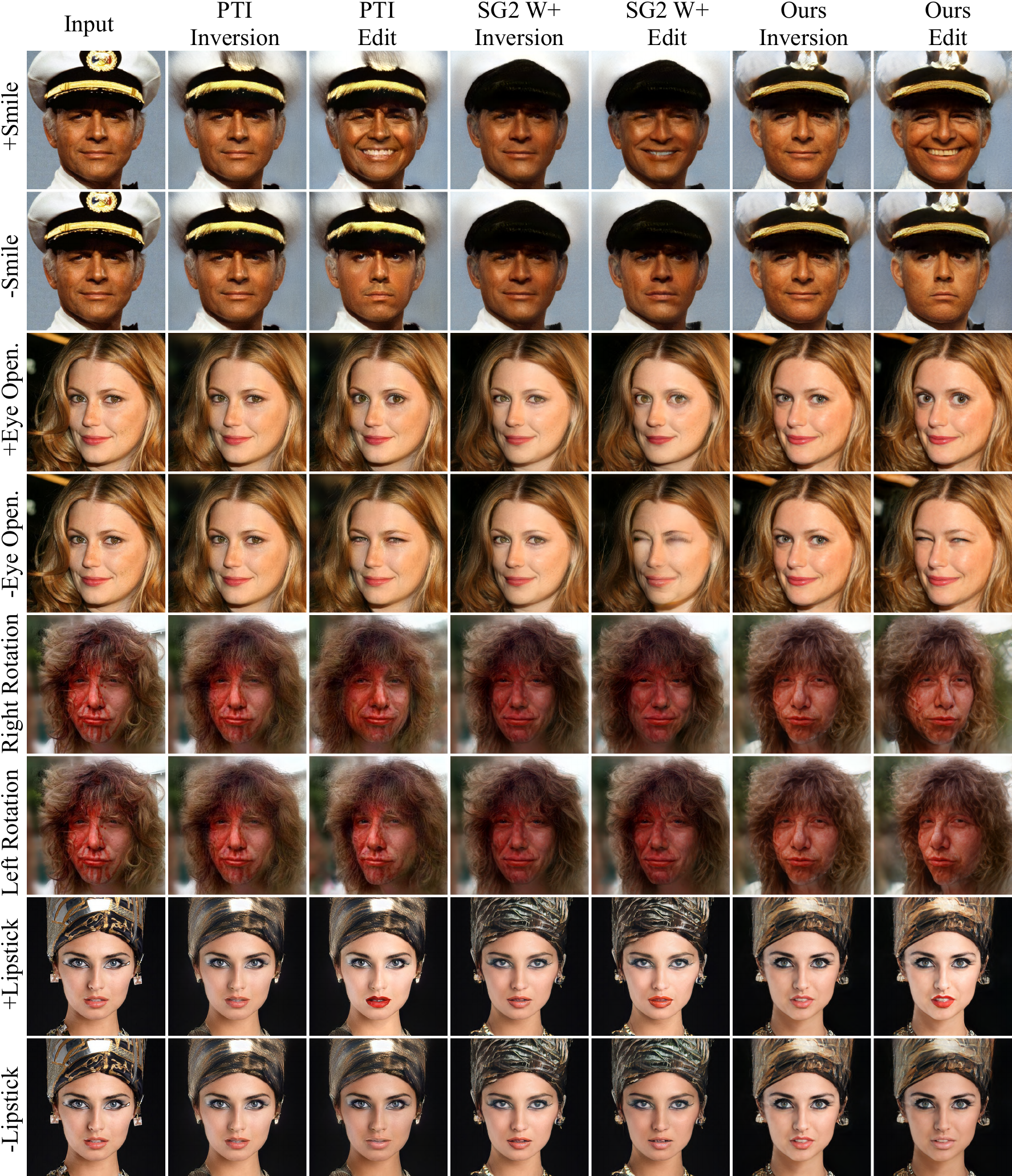}
\caption{More visual examples shown the \textbf{editability comparison} of our method compared to PTI~\cite{roich2021pivotal} and SG2 $\mathcal{W}^{+}$~\cite{abdal2019image2stylegan} on the human facial domain. Recall that such two methods require the optimization process and/or generator fine-tuning in the inference time, therefore, they run very slow. The input images are taken from the CelebA-HQ~\cite{liu2015deep, karras2018progressive} test set. The rotation and smile directions are obtained from InterFaceGAN~\cite{shen2020interpreting}, whereas other directions are borrowed from GANSpace~\cite{harkonen2020ganspace}. Best viewed in zoom.}
\label{fig:opt_based_face_edit_supp_1}
\end{figure*}

\begin{figure*}[h]
\centering
\includegraphics[width=\linewidth]{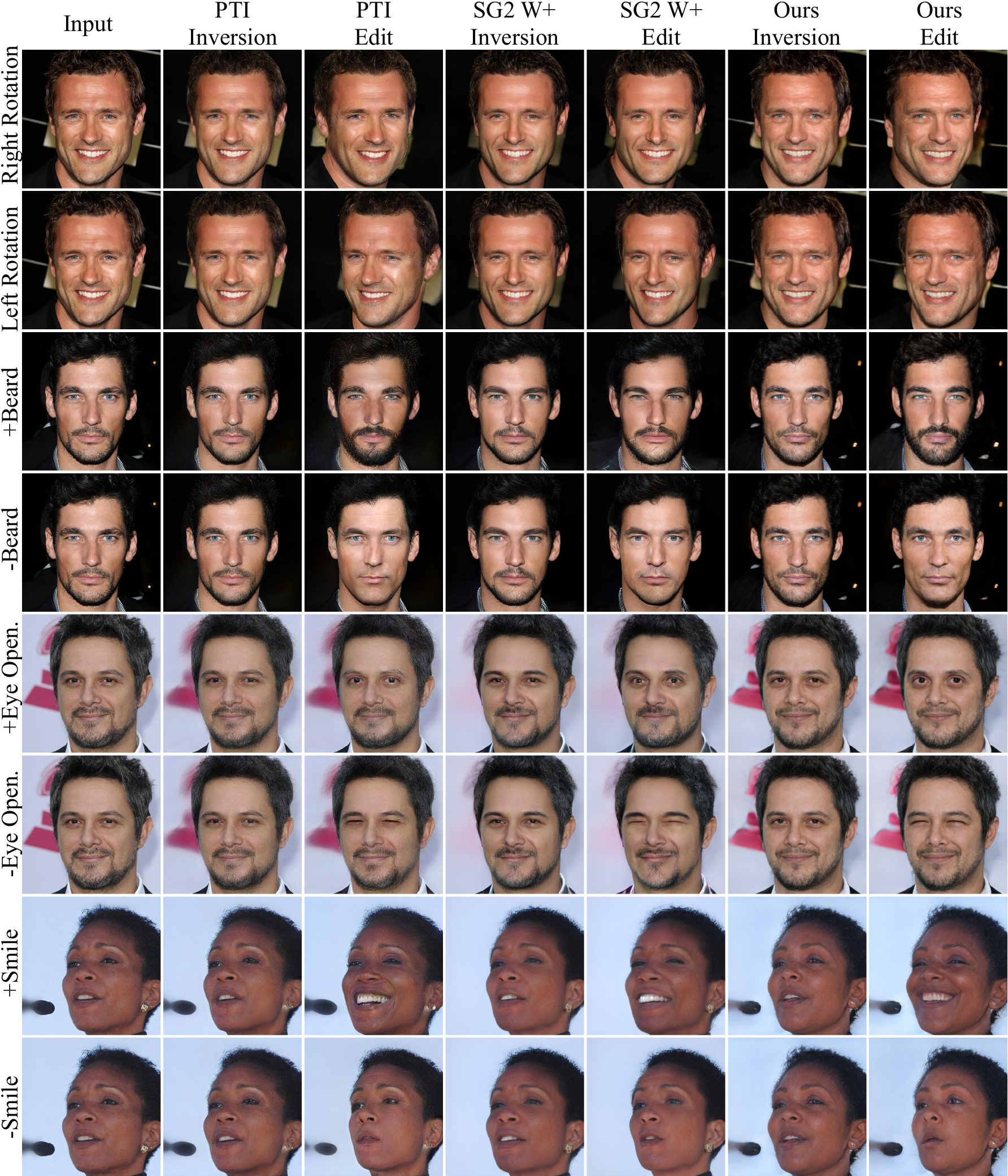}
\caption{More visual examples shown the \textbf{editability comparison} of our method compared to PTI~\cite{roich2021pivotal} and SG2 $\mathcal{W}^{+}$~\cite{abdal2019image2stylegan} on the human facial domain. Recall that such two methods require the optimization process and/or generator fine-tuning in the inference time, therefore, they run very slow. The input images are taken from the CelebA-HQ~\cite{liu2015deep, karras2018progressive} test set. The rotation and smile directions are obtained from InterFaceGAN~\cite{shen2020interpreting}, whereas other directions are borrowed from GANSpace~\cite{harkonen2020ganspace}. Best viewed in zoom.}
\label{fig:opt_based_face_edit_supp_2}
\end{figure*}

\begin{figure*}[h]
\centering
\includegraphics[width=\linewidth]{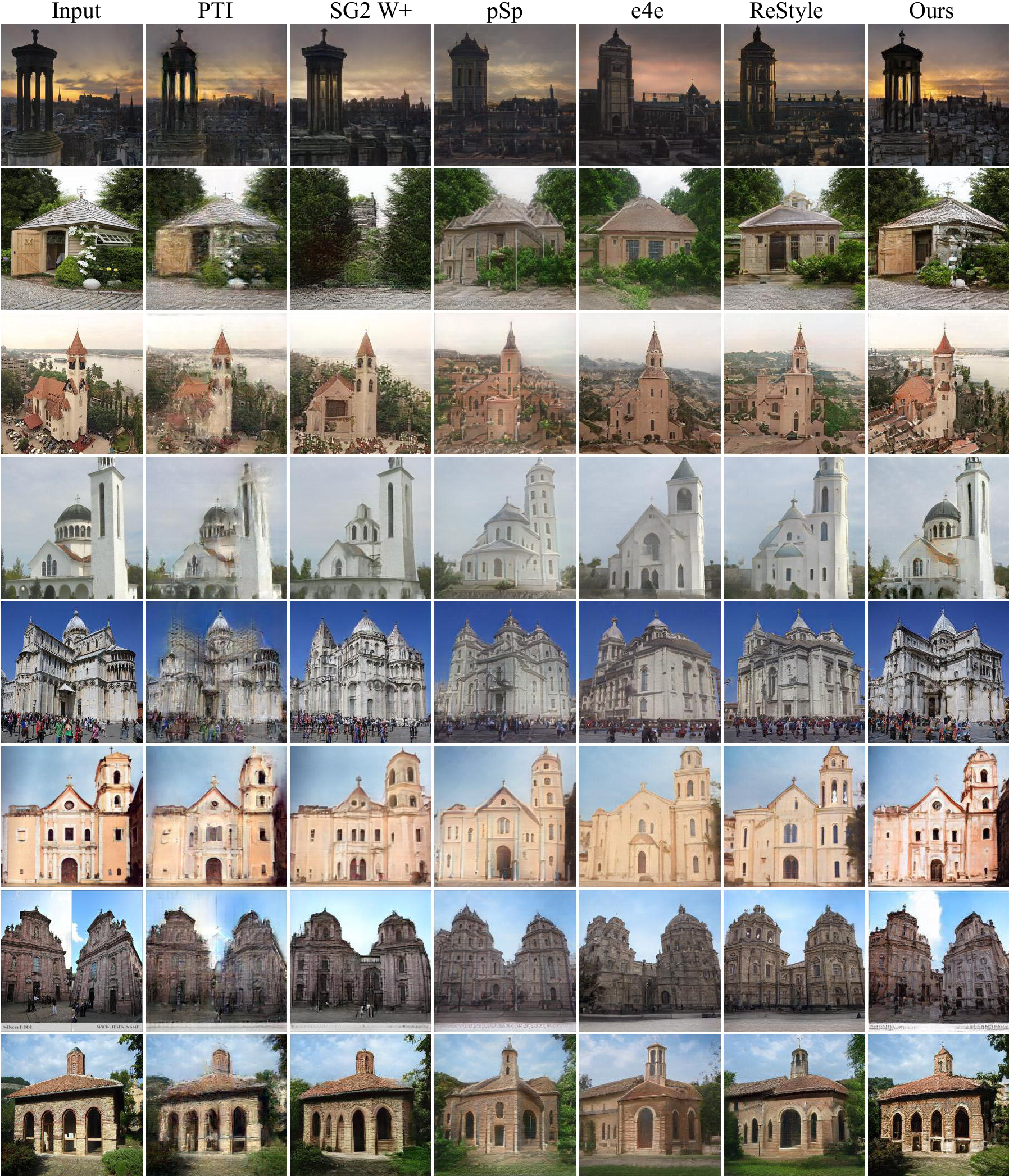}
\caption{More visual examples shown the \textbf{reconstruction comparison} of our method with other \emph{encodered-based approaches}: pSp~\cite{richardson2021encoding}, e4e~\cite{tov2021designing}, ReStyle~\cite{alaluf2021restyle}; \emph{optimization-based methods}: SG2~$\mathcal{W}^{+}$~\cite{abdal2019image2stylegan}; \emph{two-stage works}: PTI~\cite{roich2021pivotal} on the churches domain. The input images are taken from the LSUN Church~\cite{yu2015lsun} test set. Best viewed in zoom.}
\label{fig:church_reconstruction_supp_1}
\end{figure*}

\begin{figure*}[h]
\centering
\includegraphics[width=\linewidth]{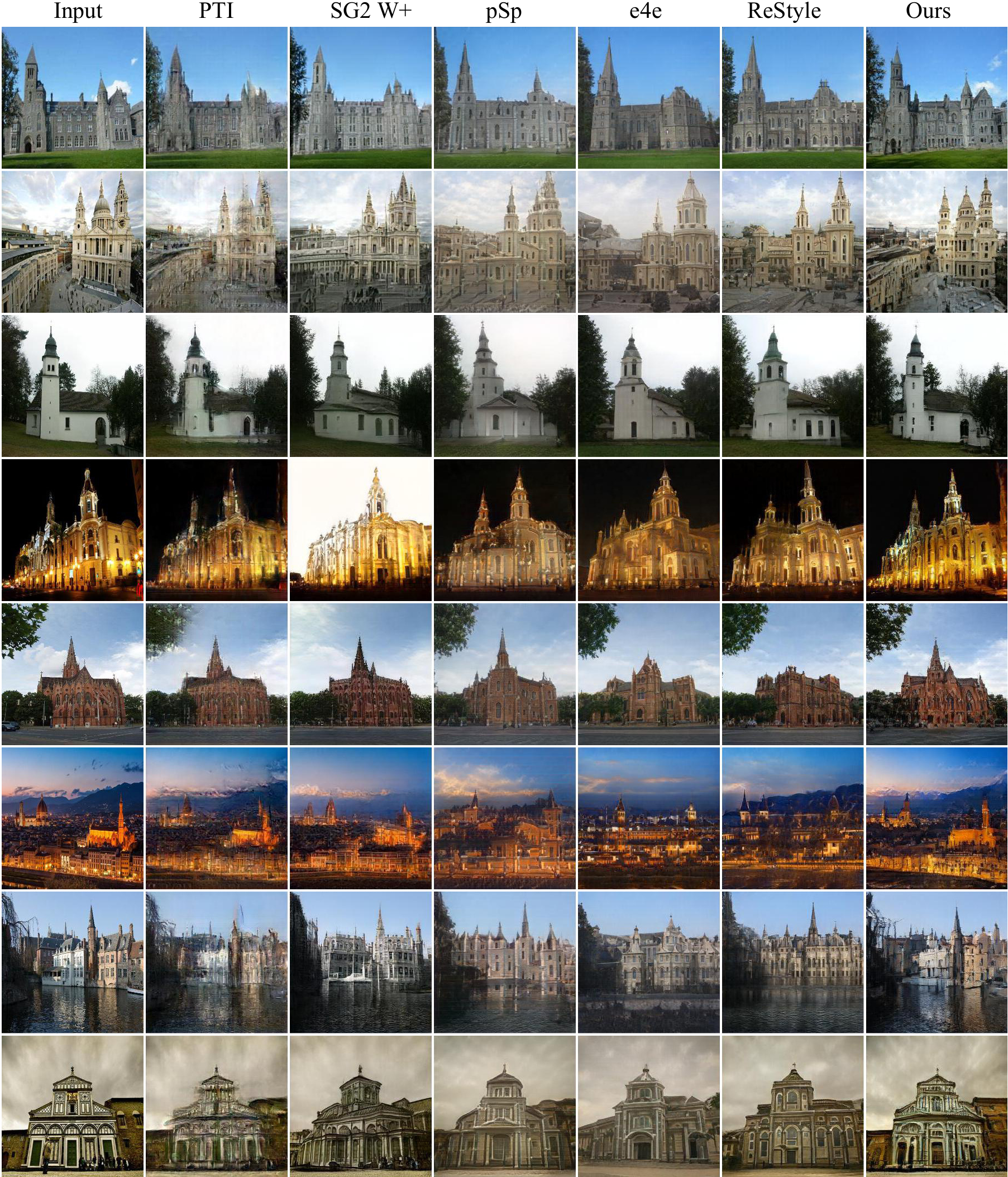}
\caption{More visual examples shown the \textbf{reconstruction comparison} of our method with other \emph{encodered-based approaches}: pSp~\cite{richardson2021encoding}, e4e~\cite{tov2021designing}, ReStyle~\cite{alaluf2021restyle}; \emph{optimization-based methods}: SG2~$\mathcal{W}^{+}$~\cite{abdal2019image2stylegan}; \emph{two-stage works}: PTI~\cite{roich2021pivotal} on the churches domain. The input images are taken from the LSUN Church~\cite{yu2015lsun} test set. Best viewed in zoom.}
\label{fig:church_reconstruction_supp_2}
\end{figure*}

\begin{figure*}[h]
\centering
\includegraphics[width=\linewidth]{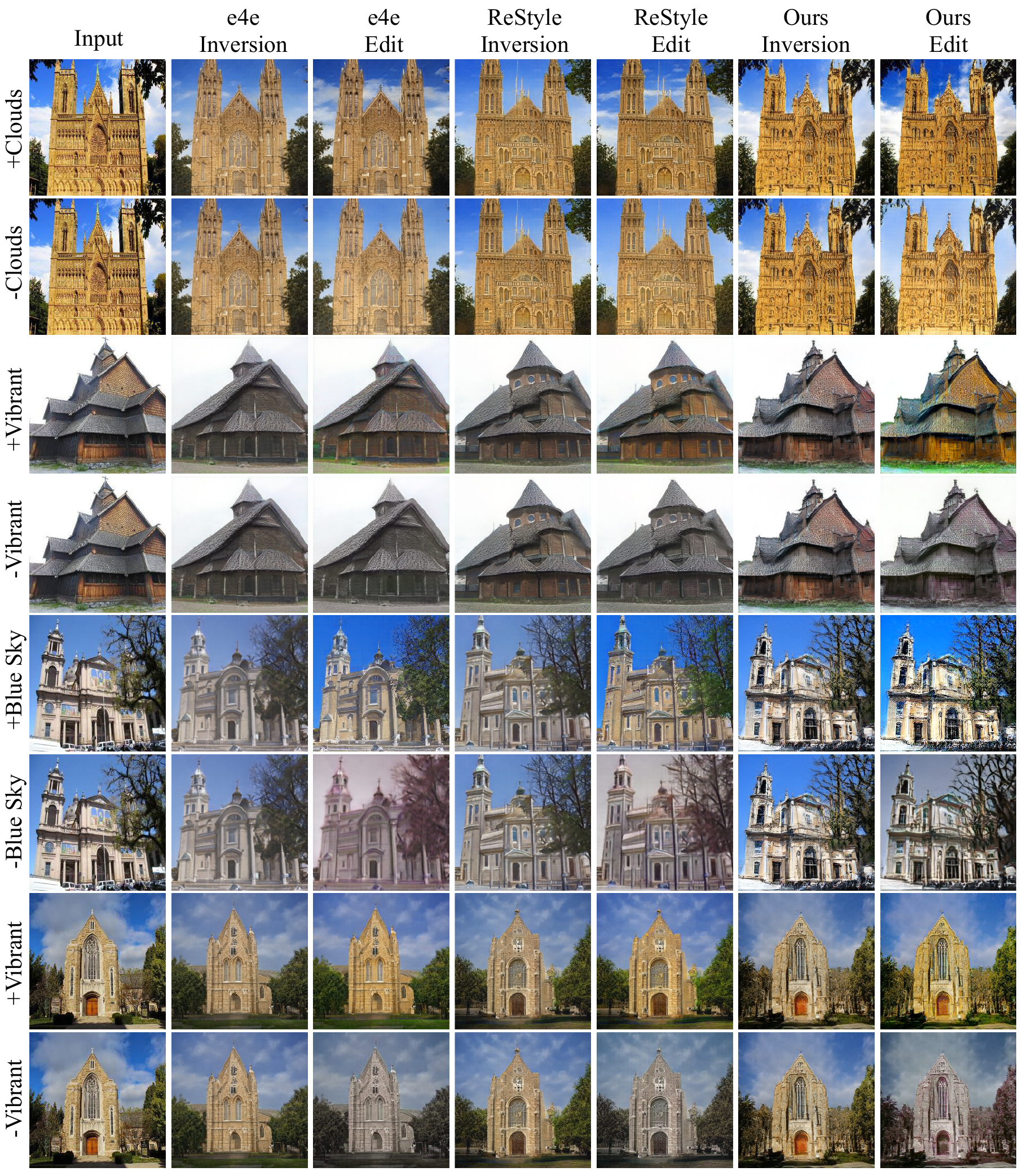}
\caption{More visual examples shown the \textbf{editability comparison} of our method with the existing \emph{encodered-based approaches}, which are e4e~\cite{tov2021designing}, ReStyle~\cite{alaluf2021restyle} on the churches domain. The input images are taken from the LSUN Church~\cite{yu2015lsun} test set. The editing directions are obtained from GANSpace~\cite{harkonen2020ganspace}. Best viewed in zoom.}
\label{fig:encoder_based_church_edit_supp_1}
\end{figure*}

\begin{figure*}[h]
\centering
\includegraphics[width=\linewidth]{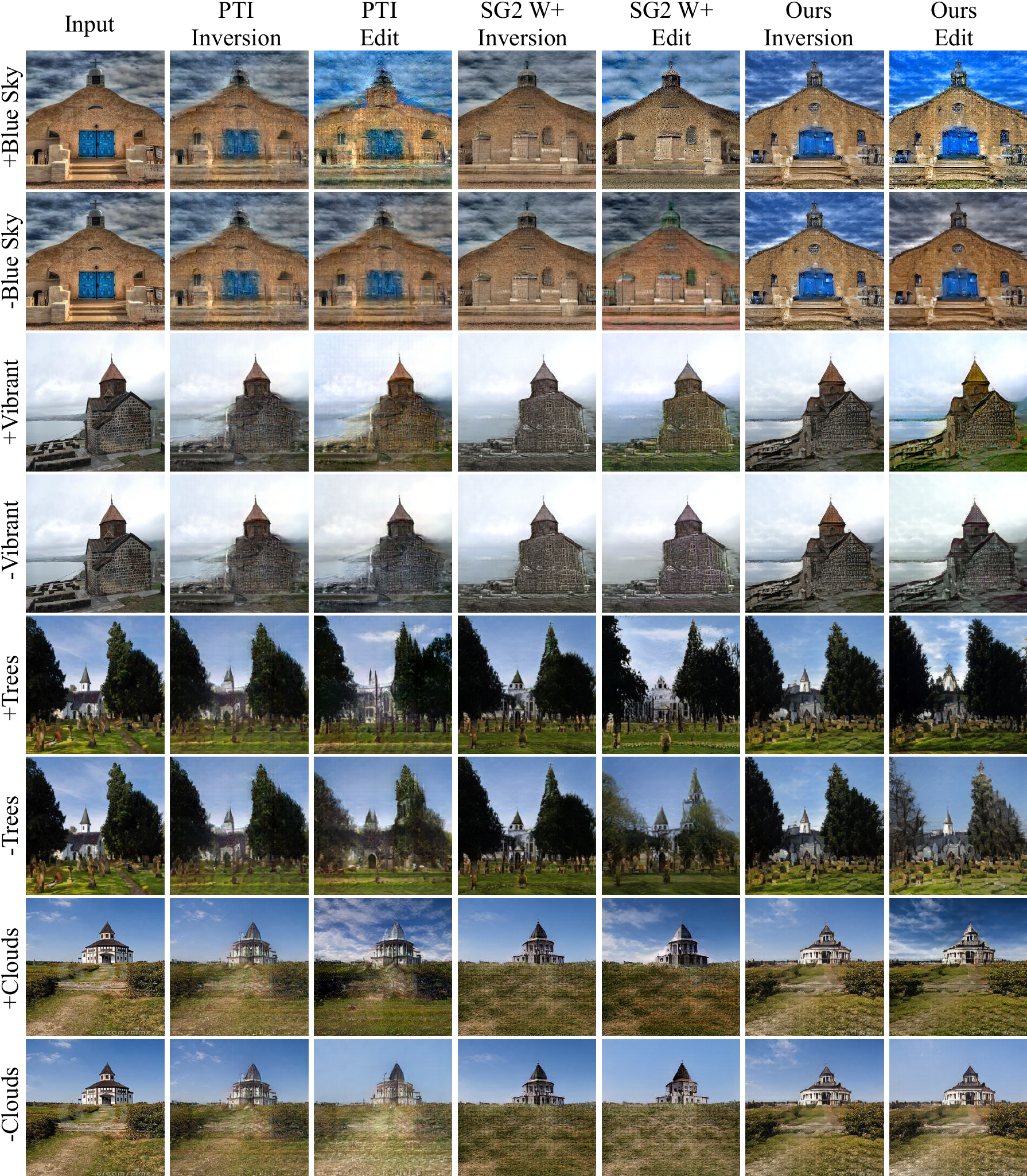}
\caption{More visual examples shown the \textbf{editability comparison} of our method compared to PTI~\cite{roich2021pivotal} and SG2 $\mathcal{W}^{+}$~\cite{abdal2019image2stylegan} on the churches domain. Recall that such two methods require the optimization process and/or generator fine-tuning in the inference time, therefore, they run very slow. The input images are taken from the LSUN Church~\cite{yu2015lsun} test set. The editing directions are obtained from GANSpace~\cite{harkonen2020ganspace}. Best viewed in zoom.}
\label{fig:opt_based_church_edit_supp_1}
\end{figure*}

\begin{figure*}[h]
\centering
\includegraphics[width=\linewidth]{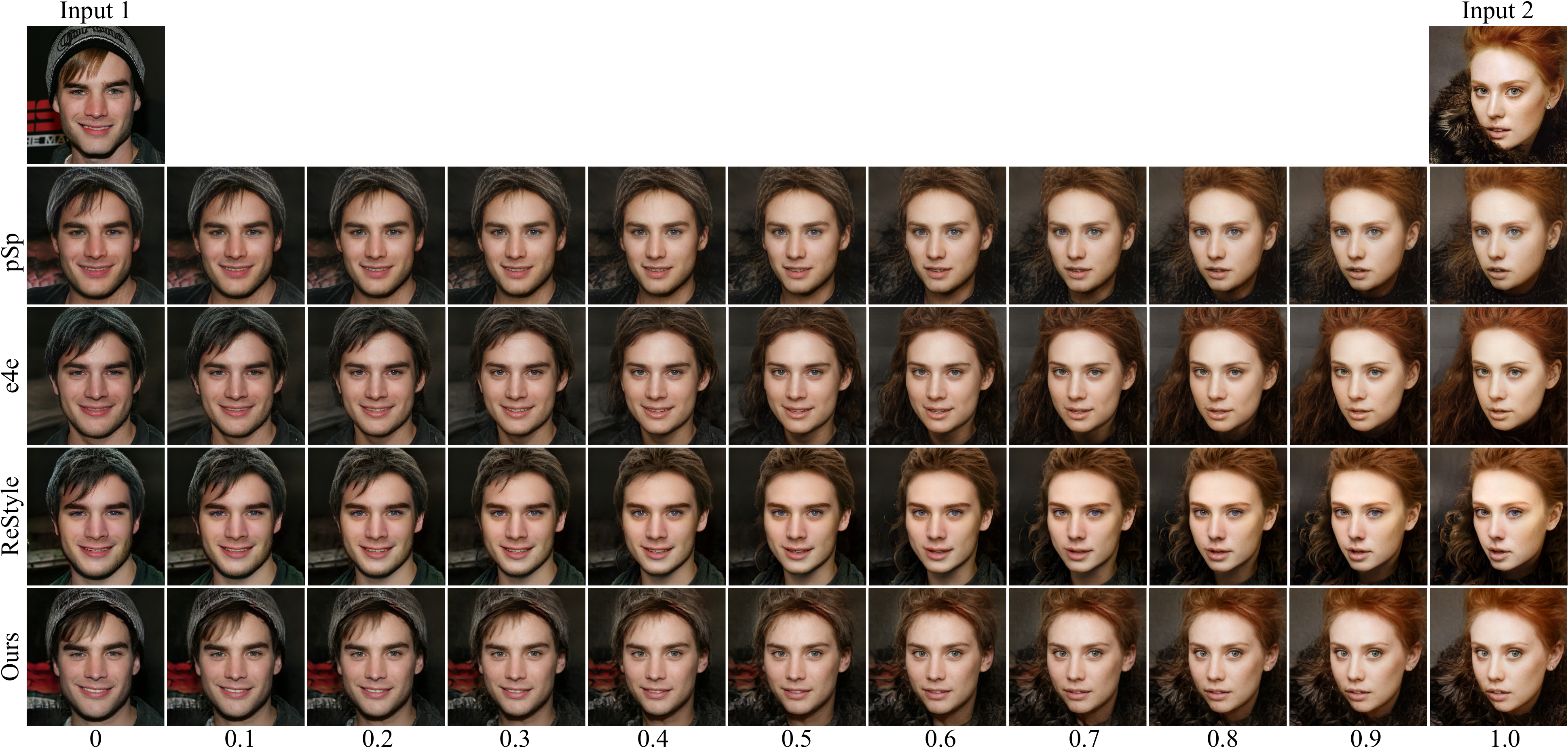}
\caption{The additional results for \textbf{real-world image interpolation} of our method compared to the existing state-of-the-art encoder-based inversion techniques. The input images are taken from the CelebA-HQ~\cite{liu2015deep, karras2018progressive} test set. Best viewed in zoom.}
\label{fig:interpolation_supp_1}
\end{figure*}

\begin{figure*}[h]
\centering
\includegraphics[width=\linewidth]{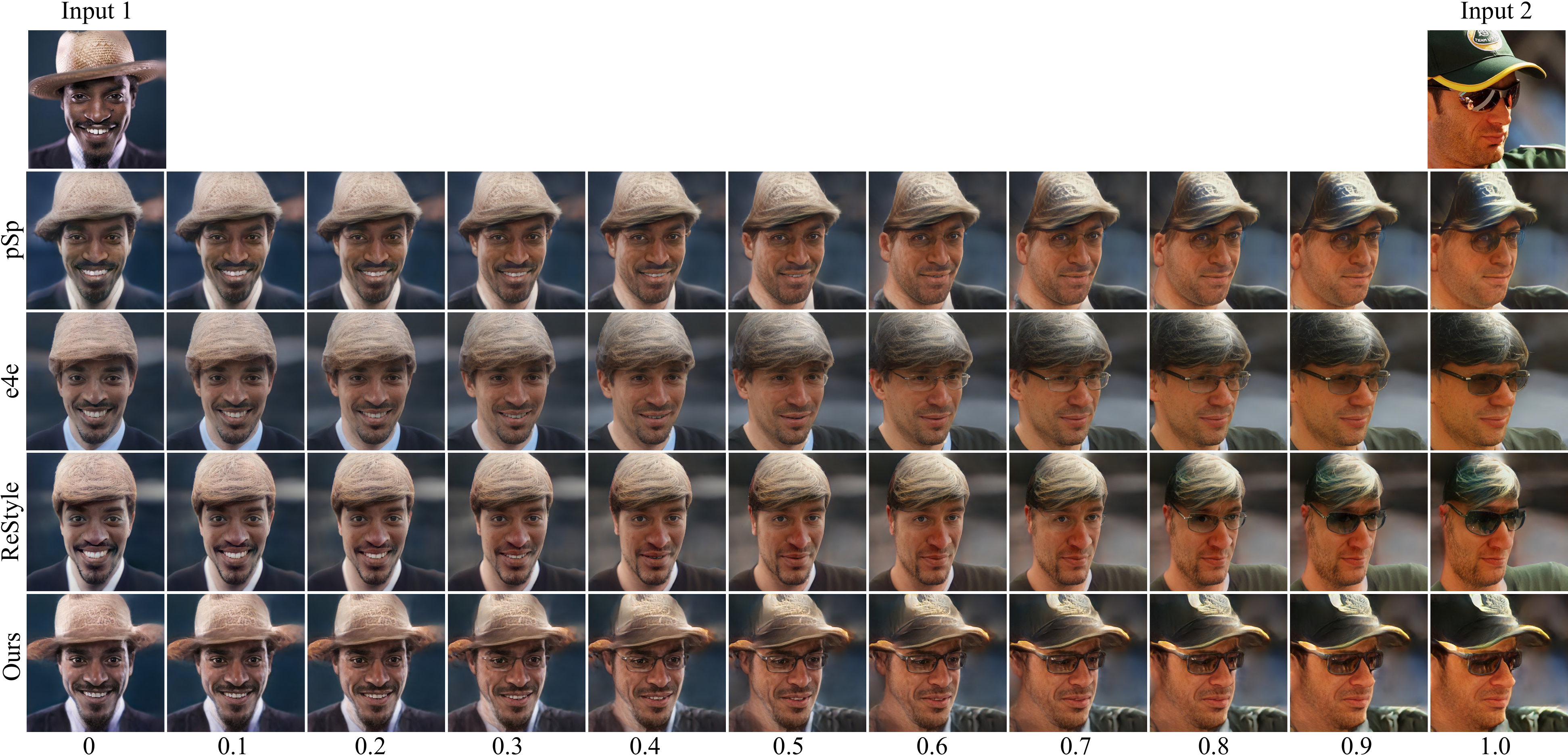}
\caption{The additional results for \textbf{real-world image interpolation} of our method compared to the existing state-of-the-art encoder-based inversion techniques. The input images are taken from the CelebA-HQ~\cite{liu2015deep, karras2018progressive} test set. Best viewed in zoom.}
\label{fig:interpolation_supp_2}
\end{figure*}

\begin{figure*}[h]
\centering
\includegraphics[width=\linewidth]{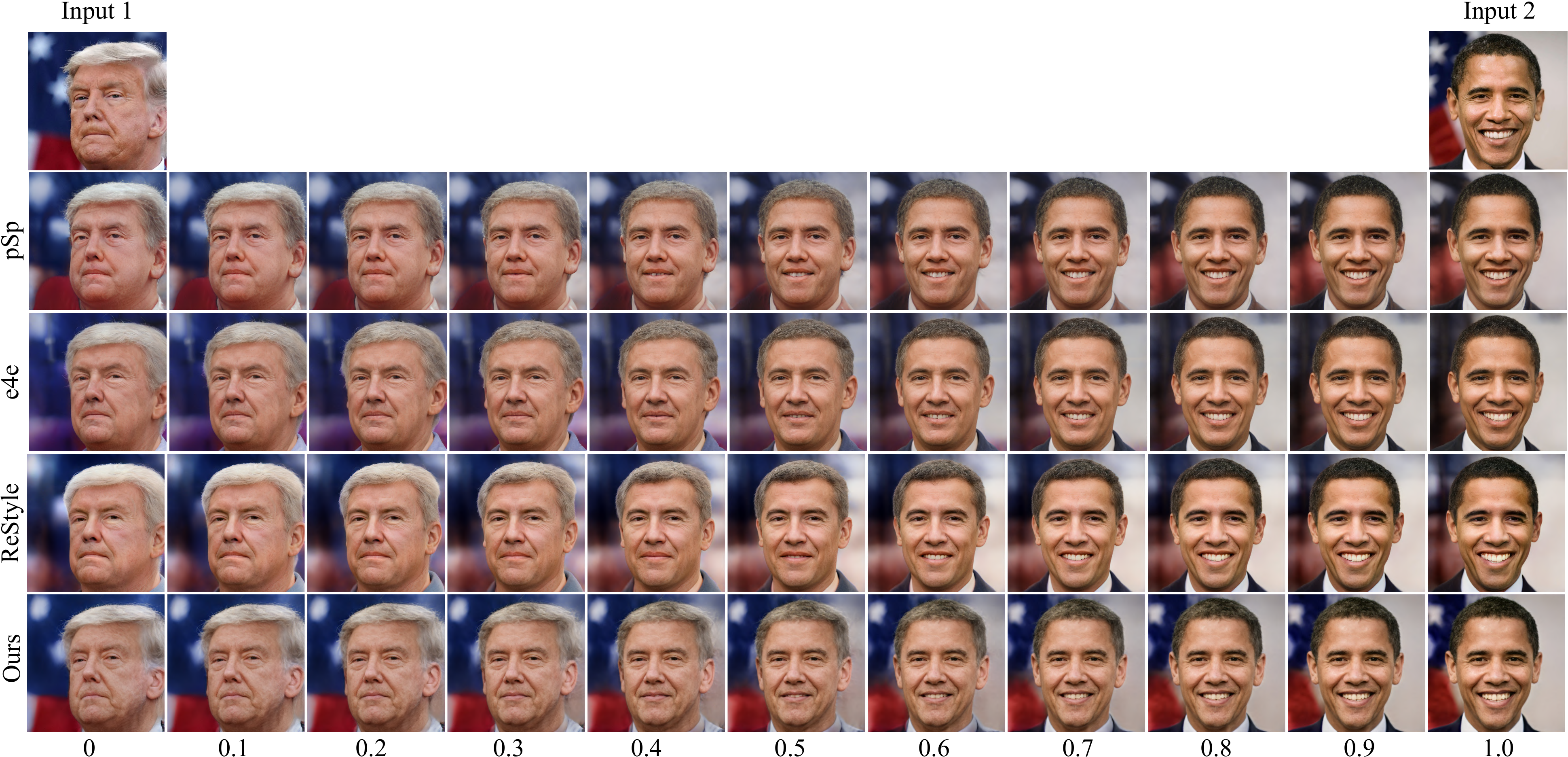}
\caption{The additional results for \textbf{real-world image interpolation} of our method compared to the existing state-of-the-art encoder-based inversion techniques. The input images are taken from the Internet. Best viewed in zoom.}
\label{fig:interpolation_supp_3}
\end{figure*}

\begin{figure*}[h]
\centering
\includegraphics[width=\linewidth]{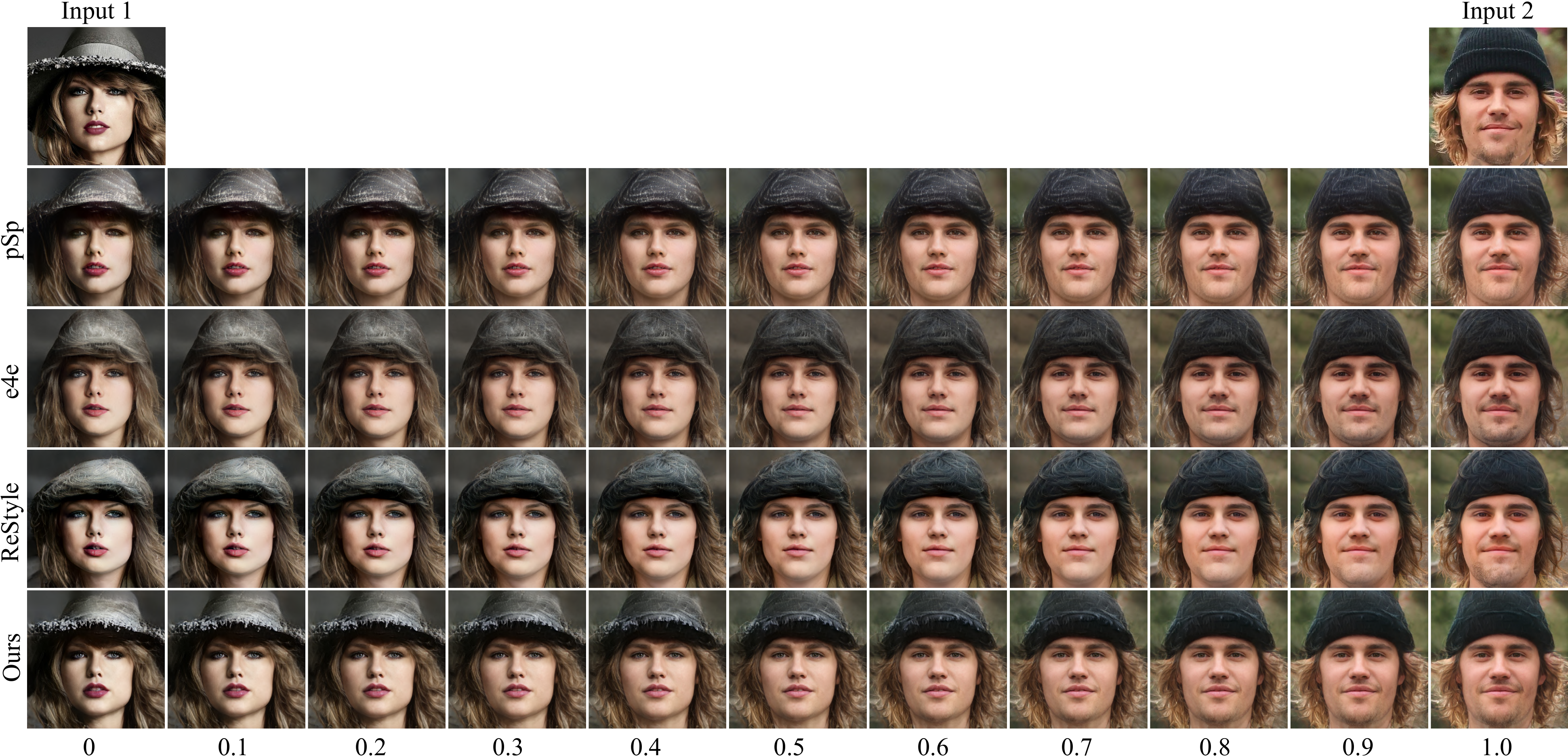}
\caption{The additional results for \textbf{real-world image interpolation} of our method compared to the existing state-of-the-art encoder-based inversion techniques. The input images are taken from the Internet. Best viewed in zoom.}
\label{fig:interpolation_supp_4}
\end{figure*}

We now turn to provide more qualitative examples on reconstruction, editability and also interpolation to further demonstrate the superiority of our method. The short descriptions for the figures are shown below. 
\begin{itemize}[leftmargin=5mm]
    \item Figure~\ref{fig:face_reconstruction_supp_1} and \ref{fig:face_reconstruction_supp_2} compare the \textbf{reconstruction} results of our method with existing state-of-the-art inversion techniques, including  encoder-based~\cite{richardson2021encoding, tov2021designing, alaluf2021restyle}, optimization-based~\cite{abdal2019image2stylegan} and two-stage methods~\cite{roich2021pivotal} for the \emph{human facial} domain on the input images taken from the CelebA-HQ~\cite{liu2015deep, karras2018progressive} test set.
    
    \item Figure~\ref{fig:encoder_based_face_edit_supp_1} and \ref{fig:encoder_based_face_edit_supp_2} compare the \textbf{editing} results of our method with the existing state-of-the-art encoder-based~\cite{richardson2021encoding, tov2021designing, alaluf2021restyle} inversion techniques for the \emph{human facial} domain on the input images taken from the CelebA-HQ~\cite{liu2015deep, karras2018progressive} test set.
    
    \item Figure~\ref{fig:opt_based_face_edit_supp_1} and \ref{fig:opt_based_face_edit_supp_2} compare the \textbf{editing} results of our method with PTI~\cite{roich2021pivotal} and SG2 $\mathcal{W}^{+}$~\cite{abdal2019image2stylegan} for the \emph{human facial} domain on the input images taken from the CelebA-HQ~\cite{liu2015deep, karras2018progressive} test set.
    
    \item Figure~\ref{fig:church_reconstruction_supp_1} and \ref{fig:church_reconstruction_supp_2} compare the \textbf{reconstruction} results of our method with existing state-of-the-art inversion techniques, including  encoder-based~\cite{richardson2021encoding, tov2021designing, alaluf2021restyle}, optimization-based~\cite{abdal2019image2stylegan} and two-stage methods~\cite{roich2021pivotal} for the \emph{churches} domain on the input images taken from the LSUN Church~\cite{yu2015lsun} test set.
    
     \item Figure~\ref{fig:encoder_based_church_edit_supp_1} compares the \textbf{editing} results of our method with the existing state-of-the-art encoder-based~\cite{richardson2021encoding, tov2021designing, alaluf2021restyle} inversion techniques for the \emph{churches} domain on the input images taken from the LSUN Church~\cite{yu2015lsun} test set.
     
     \item Figure~\ref{fig:opt_based_church_edit_supp_1} compares the \textbf{editing} results of our method with PTI~\cite{roich2021pivotal} and SG2 $\mathcal{W}^{+}$~\cite{abdal2019image2stylegan} on the \emph{churches} domain on the images taken from the LSUN Church~\cite{yu2015lsun} test set.
     
     \item Figure~\ref{fig:interpolation_supp_1}, \ref{fig:interpolation_supp_2}, \ref{fig:interpolation_supp_3}, and \ref{fig:interpolation_supp_4} show additional results for \textbf{real-world image interpolation} of our proposed pipeline, which interpolates both latent codes and generator weights compared to the common-used pipeline, which interpolates latent codes only.
\end{itemize}

It is worth noting that since the images have a very large resolution, which is $1024 \times 1024$ and $256 \times 256$ for human faces and churches, respectively, therefore, we recommend zoomed-in when viewing to have the best judgment. 

\clearpage

\end{document}